\documentclass[preprint,3p,twocolumn]{elsarticle}

\usepackage{amssymb}
\usepackage{amsthm}
\usepackage[ruled,vlined,linesnumbered]{algorithm2e} 
\usepackage{algorithmic}
\usepackage{graphicx}
\usepackage{epsfig}
\usepackage{amsmath}
\usepackage{lscape}
\usepackage{tabularx}
\usepackage{color, soul}
\usepackage{makecell}
\usepackage[table]{xcolor}
\usepackage{colortbl}
\usepackage{rotating}
\usepackage{mathtools}
\usepackage[colorlinks = true,
            linkcolor = blue,
            urlcolor  = blue,
            citecolor = blue,
            anchorcolor = blue]{hyperref}
\usepackage[left]{lineno}

\makeatletter
\def\ps@pprintTitle{%
  \let\@oddhead\@empty
  \let\@evenhead\@empty
  \let\@oddfoot\@empty
  \let\@evenfoot\@oddfoot
}
\makeatother

\renewcommand{\hl}[1]{#1}

\begin{document}

\begin{frontmatter}

\title{Dynamic Obstacle Avoidance of Unmanned Surface Vehicles in Maritime Environments Using Gaussian Processes Based Motion Planning}

\author[1,2]{Jiawei~Meng\corref{cor1}}
\ead{jiawei.meng@ucl.ac.uk}
\ead{jiaweimeng1994@gmail.com}
\author[1]{Yuanchang~Liu}
\author[2]{Danail~Stoyanov}
\cortext[cor1]{Corresponding author: Jiawei Meng}
\address[1]{Department of Mechanical Engineering, University College London, London WC1E 6BT, UK}
\address[2]{Department of Computer Science, University College London, London WC1E 6BT, UK}

\begin{abstract}
During recent years, unmanned surface vehicles are extensively utilised in a variety of maritime applications such as the exploration of unknown areas, autonomous transportation, offshore patrol and others. In such maritime applications, unmanned surface vehicles executing relevant missions that might collide with potential static obstacles such as islands and reefs and dynamic obstacles such as other moving unmanned surface vehicles. To successfully accomplish these missions, motion planning algorithms that can generate smooth and collision-free trajectories to avoid both these static and dynamic obstacles in an efficient manner are essential. In this article, we propose a novel motion planning algorithm named the Dynamic Gaussian process motion planner 2, which successfully extends the application scope of the Gaussian process motion planner 2 into complex and dynamic environments with both static and dynamic obstacles. First, we introduce an approach to generate safe areas for dynamic obstacles using modified multivariate Gaussian distributions. Second, we introduce an approach to integrate real-time status information of dynamic obstacles into the modified multivariate Gaussian distributions. The multivariate Gaussian distributions with real-time statuses of dynamic obstacles can be innovatively added into the optimisation process of factor graph to generate an optimised trajectory. \hl{We also develop a variant of the proposed algorithm that integrates the international regulations for preventing collisions at sea, enhancing its operational effectiveness in maritime environments.} The proposed algorithms have been validated in a series of benchmark simulations and a dynamic obstacle avoidance mission in a high-fidelity maritime environment in the Robotic operating system to demonstrate the functionality and practicability.
\end{abstract}


\end{frontmatter}

\section{Introduction}

 \begin{table}[t!]
 \caption{A list of abbreviations in this article.}
 \label{Table:abbr}
 \centering
 \footnotesize
 \begin{tabular}{m{2.0cm} m{4.5cm}}
 \hline
 \textbf{Abbreviations} & \textbf{Definitions} \\ \hline
 2-D & 2 dimensional\\
 3-D & 3 dimensional\\
 k-D & k dimensional\\
 AIS & Automatic identification system\\
 AUV & Autonomous underwater vehicle\\
 CV & Constant-velocity\\
 CV-SL & Constant-velocity straight-line\\
 \hl{COLREG} & \hl{Convention on the international regulations for preventing collisions at sea}\\
 D-GPMP2 & Dynamic Gaussian process motion planner 2\\
 \hl{D-GPMP2-COLREGs} & \hl{Dynamic Gaussian process motion planner 2 based on COLREG rules}\\ 
 \hl{DWA} & \hl{Dynamic window approach}\\
 GPS & Global positioning system\\
 GP & Gaussian process \\
 GP-based & Gaussian-process-based \\
 GPMP2 & Gaussian process motion planner 2\\
 LTV-SDE & Linear time-varying stochastic differential equation\\
 MAP & Maximum a posterior\\
 ROS & Robotics operating system \\ 
 USV & Unmanned surface vehicle\\
 WAM-V 20 & Wave adaptive modular vessel 20\\
 \hline
 \end{tabular}
 \end{table}
 
 \hl{Owing to the inherent flexibility and operational versatility, USVs have been extensively deployed in diverse autonomous maritime missions over recent decades.} \hl{The deployment of USVs for autonomous maritime operations comprises three essential elements:} first, sensing the surrounding maritime environments through onboard sensors, second, planning safe and smooth trajectories based on sensed information and third, navigating the platforms to follow the generated trajectories to perform relevant missions. Sensing and navigation are effectively connected through planning, which is responsible for processing sensed information as well as for generating optimised trajectories for navigation. It is important to factor in not just static obstacles such as islands and reefs, but also dynamic obstacles such as other moving USVs during the motion planning processes of USVs. \hl{Furthermore, motion planning algorithms are required to plan safe and smooth trajectories within an extreme short time such as 100 [ms] after detecting any dynamic obstacle in maritime environments. Otherwise, the USV may already have collided with dynamic obstacles during the planning latency period, even if subsequently generated trajectories are both safe and smooth.} 
 
 Current mainstream motion planning algorithms contain: 1) gird-based algorithms \cite{meng2022anisotropic, meng2022fully, dijkstra2022note, hart1968formal, petres2007path}, 2) sampling-based algorithm \cite{meng2022anisotropic, meng2022fully, kavraki1996probabilistic, lavalle1998rapidly, meng2018uav, gammell2014informed, kuffner2000rrt}, 3) potential field algorithms \cite{meng2022anisotropic, meng2022fully, khatib1986real, petres2005underwater}, 4) intelligent algorithms \cite{meng2022anisotropic, meng2022fully, colorni1991distributed, whitley1994genetic, mahmoudzadeh2016novel}, 5) GP-based algorithms \cite{mukadam2016gaussian, dong2016motion, huang2017motion, mukadam2018continuous} and others. \hl{They can also be divided into two categories: 1) path planning methods and 2) trajectory planning methods. In path planning, a geometric path is defined as a sequence of waypoints focusing solely on spatial coordinates. In trajectory planning, the geometric path is further augmented with time-dependent information. Except traditional mainstream motion planning algorithms, obstacle avoidance strategy can dynamically adjust the path or trajectory and directly generates control commands based on real-time sensor data during the USV's movement to prevent collisions with obstacles.} The GP-based algorithms have several benefits among the mainstream motion planning algorithms \cite{mukadam2016gaussian, dong2016motion, huang2017motion, mukadam2018continuous}. They can generate an optimised path at an extremely fast speed while smoothing and shortening the path during the planning process. Furthermore, they can eliminate the drawback of local minima by using the probabilistic inference \cite{meng2022anisotropic, mukadam2018continuous}. \hl{The benefits of GP-based algorithms compared with mainstream motion planning algorithms and obstacle avoidance strategies are summarised in Table} \ref{Benefits of GPMP2}.

  \begin{table}[t!]
     \caption{Benefits of GP-based algorithms compared with mainstream motion planning algorithms and obstacle avoidance strategies.}
     \label{Benefits of GPMP2}
     \centering
     \footnotesize
     \begin{tabular}{c c} \hline 
     \makecell{\textbf{Methods}} & \makecell{\textbf{Benefits}} \\ \hline 
     \makecell{GP-based algorithms} & \makecell{1. Short execution time\\2. Smooth path\\3. No local minima issue} \\ \hline 
     \end{tabular}
 \end{table}

 \hl{To ensure accuracy, the effects of both static obstacles and dynamic obstacles should be modelled in the motion planning problems of USVs.} Nearly all the mainstream motion planning algorithms have been improved and can avoid dynamic obstacles. There are many studies about using grid-based motion planning algorithms to avoid dynamic obstacles \cite{stentz1994optimal, zhu2021path, jiang2023dynamic, he2021dynamic}. For example, \cite{stentz1994optimal} has proposed an approach to improve the fundamental version of the grid-based motion planning algorithms and enable it to work in complex and dynamic environments. \cite{zhu2021path} \hl{proposes an improved D* Lite algorithm, a grid-based motion planning algorithm designed for marine robots such as USVs to navigate dynamic maritime environments while avoiding moving obstacles.} Similar to grid-based motion planning algorithms, there are also many studies about using sampling-based motion planning algorithms to avoid dynamic obstacles \cite{enevoldsen2021colregs, chiang2018colreg, ferguson2006replanning, adiyatov2017novel, aoude2013probabilistically}. For example, \cite{enevoldsen2021colregs, chiang2018colreg} \hl{have proposed different approaches to combine sampling-based motion planning algorithm with COLREG collision avoidance regulations.} Furthermore, \cite{ferguson2006replanning, adiyatov2017novel} have proposed a series of strategies for the sampling-based motion planning algorithms to avoid dynamic obstacles through 1) abandoning the minority of tree branches invalid by dynamic obstacles, 2) maintaining the rest majority of tree branches and 3) generating new collision-free tree branches. Potential field motion planning algorithms and intelligent motion planning algorithms have been researched for avoiding dynamic obstacles for many years as well \cite{falanga2020dynamic, wu2023research, sonny2023q, choi2021reinforcement}. For example, \cite{falanga2020dynamic} has proposed an approach to integrate the real-time status information of dynamic obstacles into a potential field algorithm. \cite{wu2023research} has proposed a two-layer approach that combines an intelligent algorithm and a potential field algorithm to avoid dynamic obstacles.   \cite{chen2019hybrid} \hl{employs an obstacle avoidance strategy (DWA) as a local motion planner to generate collision-free paths for USVs navigating among dynamic obstacles.}
 
 \hl{Despite some prior studies has modified some mainstream motion planning algorithms and obstacle avoidance strategies to handle dynamic obstacles, there was not many studies applying the state-of-the-art trajectory optimisation method (GP-based algorithms) to deal with the motion planning problems of avoiding static and dynamic obstacles simultaneously.} Although \cite{qing2023dynamic} has investigated how to use the GP-based algorithms to avoid dynamic obstacles by proposing a new factor graph structure. But this is the only published research about this topic and it just adjusted the factor graph structure without providing a specific method to model the dynamic obstacles. To bridge this research gap, this article has specifically focused on developing a novel motion planning algorithm for USVs to avoid both static and dynamic obstacles in maritime environments. 

  In the self-constructed simulation environments in MATLAB, we have tested and validated the performance of the proposed motion planning algorithms in a series of qualitative and quantitative benchmark simulations including 1) a benchmark with different weights of dynamic obstacle likelihood, 2) a benchmark with a dynamic obstacle with different velocities, 3) a benchmark with dynamic obstacles with different areas, 4) a benchmark with multiple dynamic obstacles, 5) a benchmark with static and dynamic obstacles, \hl{6) a benchmark with a current existing obstacle avoidance method for quantitative comparison and 7) a COLREGs-compliant benchmark simulation.}
  
  The results demonstrate the following:
  
  \begin{itemize}
      \item The proposed D-GPMP2 algorithm can generate a safe and smooth trajectory to avoid dynamic obstacles by considering their areas, directions, speeds and positions. 
      \item The proposed D-GPMP2 algorithm can generate an optimised trajectory in an extremely short execution time.
      \item The proposed D-GPMP2 algorithm can be utilised in a variety of complex environments with multiple obstacles and different statuses.
      \item \hl{The proposed D-GPMP2-COLREGs algorithm effectively performs dynamic obstacle avoidance in maritime environments while maintaining compliance with COLREGs regulations.}
  \end{itemize}

 \begin{figure}[t!]
 \centering
 \includegraphics[width=1\linewidth]{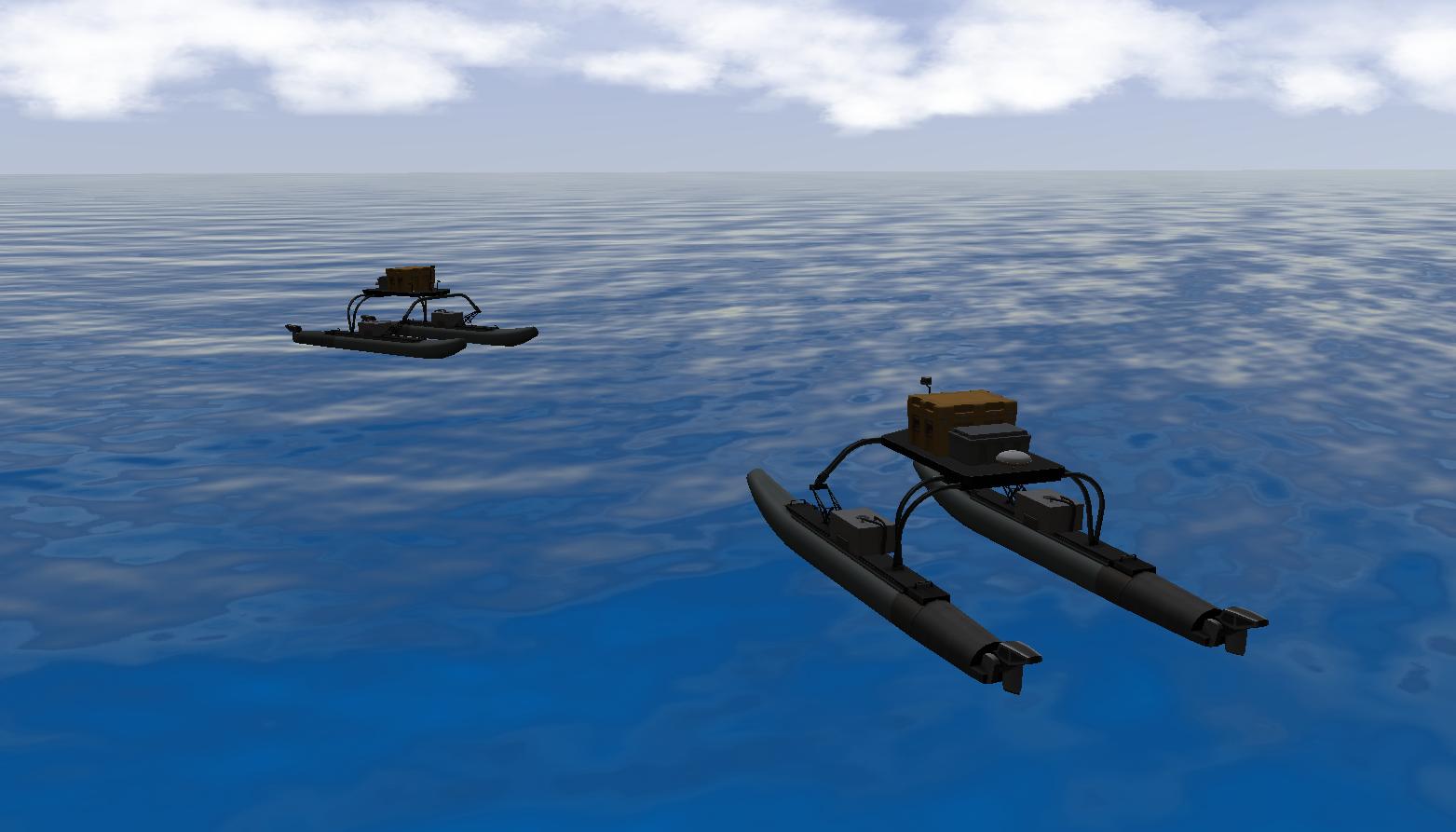}
 \centering
 \caption{Scenario demonstration in the Gazebo simulation environment in ROS: two USVs are approaching each other and at least one of them requires to take action to avoid the other. The high-fidelity maritime environment is provided by \cite{bingham2019toward}.}
 \label{USVs} 
 \end{figure}
  
  We have also tested and validated the proposed D-GPMP2 algorithm in a high-fidelity maritime environment in ROS. In ROS, augmented practicability is achieved based on the consideration of a series of physical properties such as hydrodynamic force, gravity and buoyancy \cite{meng2022anisotropic, bingham2019toward}. In comparison with the expensive testing process using a full-sized physical USV, validating the developed motion planning algorithm in ROS can effectively reduce the testing cost and maintain reliable test results simultaneously \cite{meng2022anisotropic, meng2022fully}. A scenario demonstration in the Gazebo simulation environment in ROS is provided in Fig. \ref{USVs}. 

  The reminder of this article is organised as follows. Section \ref{GPMP2} presents the mathematical model of the GPMP2 algorithm in detail. Section \ref{D-GPMP2} presents the proposed D-GPMP2 algorithm in detail. Section \ref{D-GPMP2-COLREGs} presents the proposed D-GPMP2-COLREGs algorithm in detail. Section \ref{Factor graph} presents a new structure of the optimisation tool used in the research. Section \ref{Simulations and discussions} demonstrates a series of simulation results of the proposed motion planning algorithms in MATLAB. Section \ref{Implementation in ROS} demonstrates the practical performance of the proposed D-GPMP2 algorithm in ROS, while Section \ref{Conclusion} concludes this article and proposes some future research directions.

\section{GPMP2}
\label{GPMP2}
 This section presents the traditional GP-bsaed motion planning algorithm (GPMP2 algorithm) and summarises the previous work in \cite{meng2022anisotropic, meng2022fully, mukadam2016gaussian, dong2016motion, huang2017motion, mukadam2018continuous}. In general, the GPMP2 algorithm considers a motion planning problem as a standard optimisation problem. To be more specific, the GPMP2 algorithm views the potential sample trajectories as a GP prior, a conditional constraint that facilitates the capability of avoiding static obstacles as a static obstacle likelihood and the optimisation process as a MAP estimation to obtain an optimal solution \cite{meng2022anisotropic, mukadam2018continuous}. The trajectories generated by GPMP2 are 1) relatively short (as GPMP2 incrementally optimise the posterior during the motion planning process) and 2) smooth (as the trajectories are time-continuous and differentiable).

\subsection{GP prior}
\label{GP prior}
 The GP prior can generate a large amount of continuous-time sample trajectories, which has a high possibility of containing one or multiple trajectories that satisfied the predefined requirements. Continuous-time trajectories are assumed to be sampled from a continuous-time GP \cite{mukadam2016gaussian, dong2016motion, mukadam2018continuous}:
 \begin{equation}
      \theta(t)\sim\mathcal{GP}(\mu(t),K(t,t')), 
 \end{equation}
 \noindent where $\mu(t)$ is a vector-valued mean function $\mu = [\mu_1 ... \mu_N]^T$ and $K(t,t')$ is a matrix-valued covariance function $K = [K(t_i, t_j)]|_{(0 \leq i , j \leq N)}$. 
 
 In the GPMP2 algorithm, this continuous-time GP is considered to be generated by a LTV-SDE. To be more specific, the following equation is used to describe the LTV-SDE: 
 \begin{equation}
      \dot{\theta}(t) = A(t)\theta(t) + u(t) + F(t)w(t), 
 \end{equation}
 \noindent where $u(t)$ is a known exogenous input, $A(t)$ and $F(t)$ are system matrices and $w(t)$ represents the white process noises. The white process noises is a GP with zero mean:
 \begin{equation}
      w(t)\sim GP(\textit{0}, Q_{c}\delta(t - t')),
 \end{equation}
 \noindent where $Q_{c}$ is a positive semi-definite time-varying power-spectral density matrix and $\delta(t - t')$ is the Dirac delta function.

 The solution of the continuous-time GP can be obtained by the process described in \cite{mukadam2018continuous, barfoot2014batch} as:
 \begin{equation}
    u(t) = \Phi(t,t_{0})u_{0} + \int_{t_{0}}^{t}\Phi(t,s)u(s)\textit{ds}
 \end{equation}
 \begin{equation}
    \begin{gathered}
        K(t, t') = \Phi(t,t_{0})K_{0}\Phi(t',t_{0})^T +\\ \int_{t_{0}}^{min(t,t')}\Phi(t,s)F(s)Q_{c}F(s)^T\Phi(t,s)^T\textit{ds},
    \end{gathered}
 \end{equation}
 \noindent where $u_{0}$ and $K_{0}$ are initial mean and covariance of the first state, respectively. $\Phi(t,s)$ is the state transition matrix from time \textit{t} to time \textit{s}.

 As a result, the GP prior distribution can be expressed by the following equation:
 \begin{equation}
    p(\theta) \propto exp\{ -\frac{1}{2}||\theta - u||^{2}_{K}\}.
 \label{prior_distribution}
 \end{equation}

 Sample trajectories from the GP prior provide the user with a large number of trajectory options from which to choose the optimised one depending on their requirements specified in the likelihoods.

 \subsection{Static obstacle likelihood}
 \label{Static obstacle likelihood}
 In the GPMP2 algorithm, a static obstacle likelihood that facilitates the event of avoiding static obstacles $l(\theta; \textit{$e_{st\_obs}$})$ is used and can be defined as a distribution in the exponential family as below:
 \begin{equation}
    l(\theta; \textit{$e_{st\_obs}$}) = exp\{ -\frac{1}{2}||h_{st\_obs}(\theta)||^{2}_{\sigma_{st\_obs}}\},
 \end{equation}
 \noindent where $e_{st\_obs}$ represents the event of avoiding static obstacles, the definition of matrix $\sigma_{\textit{st\_obs}}$ can be found in \cite{dong2016motion, mukadam2018continuous} and $h_{st\_obs}(\theta)$ is a vector-valued cost function:
 \begin{equation}
         h_{st\_obs}(\theta_{i}) = [c(d(\theta_{i}))],
 \label{static_obstacle_likelihood}
 \end{equation}
 where $c$ is the hinge loss function defined by the equation: 
 \begin{equation}
    c(d) = 
    \left\{
    \begin{array}{cr}
             -d + \epsilon, & d \leq \epsilon \\
             0, & d > \epsilon \\ 
    \end{array}
    \right.,
 \label{hinge_loss_function}
 \end{equation}
\noindent where $d$ is the signed distance of a point and $\epsilon$ is the safety distance. In general, the safety distance $\epsilon$ is defined based on user experience and the signed distance function $d$ is stated as:
 \begin{align}
         d = D - \overline{D},
 \label{signed_distance_field_function}
 \end{align}
 \noindent where $D$ is the Euclidean distance transforms function (also named as distance field). $\overline{D}$ is the complement of distance field $D$. 

 Based on the specific requirements for achieving different purposes, GPMP2 can incorporate various likelihoods. In this article, Section \ref{D-GPMP2} combines the GPMP2 algorithm with a dynamic obstacle likelihood to allow it to avoid both static and dynamic obstacles.
 
 \subsection{MAP estimation}
 The MAP estimation is an estimate of an unknown quantity, which equals the mode of the posterior distribution \cite{bassett2019maximum}. Now it is possible to perform a Bayesian inference to obtain an optimised trajectory using the knowledge provided in Sections \ref{GP prior} and \ref{Static obstacle likelihood}: 1) a GP prior that includes a variety of sample trajectories from a start point to a goal point and 2) a likelihood that facilitates the event of avoiding static obstacles. Thus, the Bayesian inference of this problem can be defined as:   
 \begin{equation}
 \begin{gathered}
    \theta_{0:N}^{*}=\mathop{\arg\max}_{\theta}p(\theta|{e}) = \mathop{\arg\max}_{\theta} \; p(\theta)p(e_{st\_obs}|\theta).
 \end{gathered}
 \label{MAP1}
 \end{equation}

 \begin{figure}[t!]
 \centering
 \includegraphics[width=1 \linewidth]{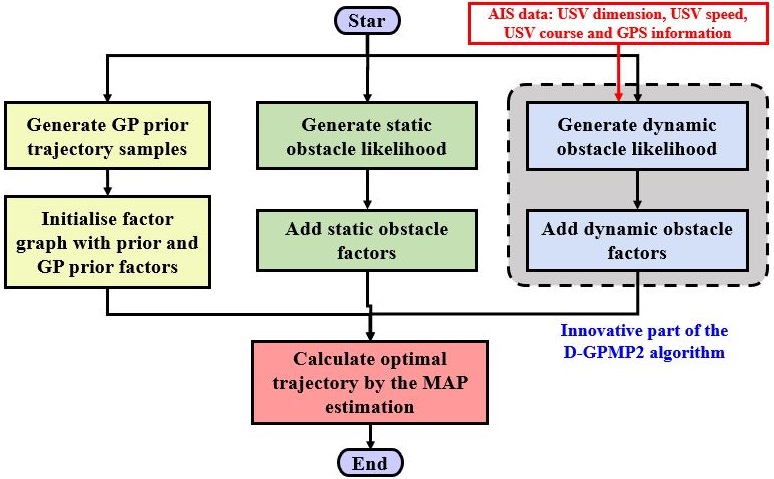}
 \centering
 \caption{Flow diagram of the proposed D-GPMP2 algorithm. The innovative part enhances the original GPMP2 algorithm into the proposed D-GPMP2 algorithm that can avoid static and dynamic obstacles simultaneously.}
 \label{overall_algorithm_dgpmp2} 
 \end{figure}
 
 The MAP estimation of a GP can be converted into a least squares problem as described in \cite{mukadam2018continuous}, which has been well-studied for many years and there are many methods to solve least squares problem including but not limited to Gaussian-Newton method and Levenberg–Marquardt method \cite{gauss1877theoria, levenberg1944method}. In this work, Levenberg–Marquardt method is used to solve the least squares problem in any GP-based motion planning algorithm.

\section{D-GPMP2}
\label{D-GPMP2}
 This section presents the proposed D-GPMP2 algorithm, in detail, which includes two new parts added into the original GPMP2 algorithm: 1) dynamic obstacle area modelling part and 2) dynamic obstacle velocity modelling part. Compared with previous studies of using multivariate Gaussian distributions to model dynamic obstacles, the novelties of our method are: 1) a specific way is proposed to integrate sensor data such as AIS data, camera data and other data into the dynamic obstacle modelling and 2) our method consider both the area and velocity of a dynamic obstacle during the modelling process.

 \begin{figure}[t!]
 \centering
 \includegraphics[width=1 \linewidth]{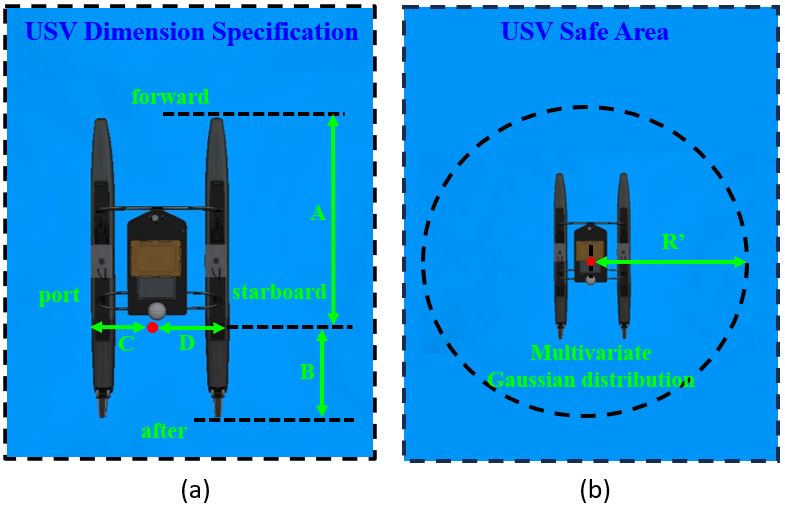}
 \centering
 \caption{(a) USV dimension specification from AIS data. (b) USV safe area based on the multivariate Gaussian distribution radius $R'$ and $R'$ is proportional to the self-defined safe radius $R$. The USV model in the above sub-figures is from \cite{bingham2019toward}.}
 \label{AIS_data_safe_range} 
 \end{figure}
 
 For the dynamic obstacle area modelling, we propose a method that combines the sensor data such as AIS data, camera data and other data and multivariate Gaussian distribution to construct a safe area for the target USV. 
 
 For the dynamic obstacle velocity modelling, we propose: 1) a method which can integrate the speed of the target USV into the multivariate Gaussian distribution and 2) a function that can rotate the multivariate Gaussian distribution with speed information to the correct direction and translate the rotated multivariate Gaussian distribution to the correct location. 
 
 Based on these two new parts, a dynamic obstacle likelihood can be constructed and the MAP estimation of the proposed D-GPMP2 algorithm considers both the static obstacle likelihood and the dynamic obstacle likelihood. By adding dynamic obstacle likelihood, the proposed D-GPMP2 algorithm gives the original GPMP2 algorithm the capability of avoiding static and dynamic obstacles. The flow diagram of the proposed D-GPMP2 algorithm is illustrated in Fig. \ref{overall_algorithm_dgpmp2}.
 
\subsection{Dynamic obstacle area modelling}
\label{Dynamic Obstacle Area Modelling}
 \hl{Dynamic obstacle area modelling method used in this article automatically adjusts model boundaries to reflect real-world obstacle sizes.} \hl{Different sensors can be installed on a USV to collect information about nearby obstacles.} For example, the AIS is a fundamental equipment mounted on the USVs to display maritime traffic information and communicate with other USVs. \hl{According to the official website of the USV platform used in this article, the WAM-V USV platform offers factory-configured sensors, such as AIS and other options} \cite{opt2025}. In general, maritime traffic information information from the AIS contains: 1) the target USV position from the GPS, 2) the target USV dimension, 3) the target USV speed, 4) the target USV course (also known as target USV direction) and others \cite{xiao2015comparison}. \hl{When the AIS is not functioning properly, alternative sensors such as cameras, event cameras and GPS can be utilised to obtain similar information.}

 \begin{figure}[t!]
 \centering
 \includegraphics[width=1 \linewidth]{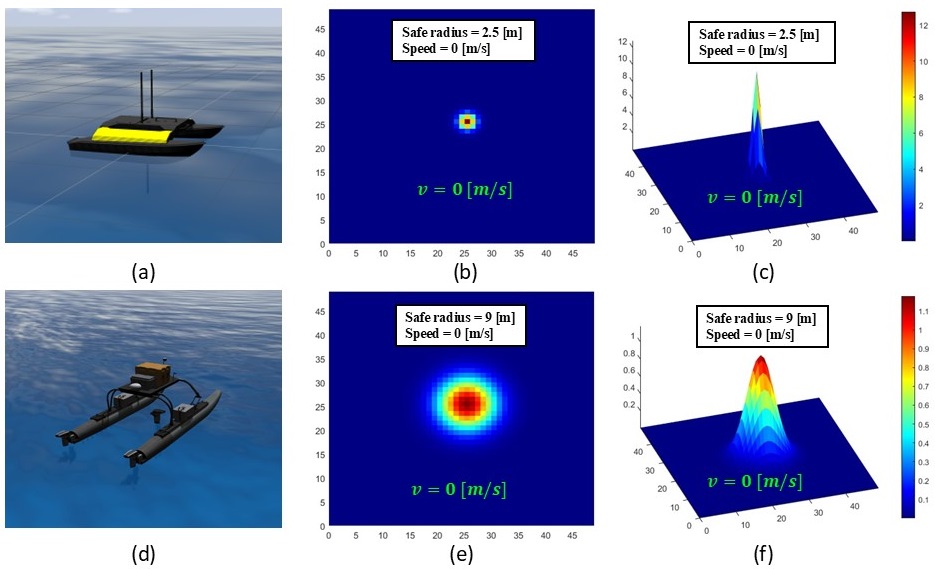}
 \centering
 \caption{(a) Heron USV in virtual environment \cite{andrade2020development}. (b) The safe area of the Heron USV from the 2-D perspective. (c) The safe area of the Heron USV from the 3-D perspective.  (c) WAM-V 20 USV in virtual environment \cite{bingham2019toward}. (d) The safe area of the WAM-V 20 USV from the 2-D perspective. (e) The safe area of the WAM-V 20 USV from the 3-D perspective. In the colour bars, a larger value indicates a higher risk and a smaller value indicates a lower risk.}
 \label{2D_Gaussian_different_areas} 
 \end{figure}

 \begin{table}[t!]
 \caption{USV dimension specification \cite{meng2022fully, heron}.}
 \label{Table:vessels}
 \centering
 \footnotesize
 \begin{tabular}{c c c}
 \hline
 \textbf{USV Name} & \textbf{USV Length} & \textbf{USV Width} \\ \hline
 WAM-V 20 USV & 6 [m] & 3 [m] \\ 
 Heron USV & 1.35 [m] & 0.98 [m] \\ \hline
 \end{tabular}
 \end{table}
 
 In maritime environments, other moving USVs are the most common dynamic obstacles for a USV. Fig. \ref{AIS_data_safe_range} (a) illustrates the information from the AIS that describes the dimension specification of a target USV: $A$ represents the distance from forward perpendicular, $B$ represents the distance from after perpendicular, $C$ represents the distance inboard from port side and $D$ represents the distance inboard from starboard side. In the meantime, Fig. \ref{AIS_data_safe_range} (b) demonstrates the USV safe area determined by a self-defined safe radius $R$ of the USV using the equation below:
 \begin{equation}
    R = \zeta \cdot \frac{\overbrace{A + B}^{length} + \overbrace{C + D}^{width}}{2},
 \label{safe_radius}   
 \end{equation}
 \noindent where $\zeta$ is an adjustable positive scaling factor \hl{and it is used for experimental calibration}, $[A+B]$ represents the length of the USV, $[C+D]$ represents the width of the USV and the USV safe area is constructed by a multivariate Gaussian distribution. Furthermore, the self-defined safe radius $R$ should ensure the multivariate Gaussian distribution at least fully covers the USV.
 
 Generally, the k-D Gaussian distribution can be written as \cite{jwo2023geometric}:
 \begin{equation}
    X\sim\mathcal{N}_{k}(\mu,\Sigma),
 \label{normal_distribution}   
 \end{equation}
 \noindent where $\mu$ is the mean vector and $\Sigma$ is the covariance matrix. The probability density function of the k-D Gaussian distribution in (\ref{normal_distribution}) can be expressed as:
 \begin{equation}
 \small
     f_{k}(X_{1},...,X_{k}) = \frac{1}{\sqrt{(2\pi)^{k}|\Sigma|}}\exp[-\frac{1}{2}(X - \mu)^T\Sigma^{-1}(X - \mu)].
 \label{pdf_kD_Gaussian}
 \end{equation}

 Due to the motion planning of USVs lies in the 2-D environments, therefore the 2-D Gaussian distribution is employed. In the 2-D situation, the probability density function can be transformed into the following form:
 \begin{equation}
 \begin{split}
     f_{2}(X, Y) = \frac{1}{2\pi\sigma_{X}\sigma_{Y}\sqrt{1 - \rho^2}}\exp\{-\frac{1}{2(1 - \rho^2)}\\ [\frac{(X - \mu_{X})^2}{\sigma_{X}^2} + \frac{(Y - \mu_{Y})^2}{\sigma_{Y}^2} - \frac{2\rho(X - \mu_{X})(Y - \mu_{Y})}{\sigma_{X}\sigma_{Y}}]\},
 \end{split}
 \end{equation}
 \noindent where $\rho$ is the correlation coefficient of $X$ and $Y$. The mean vector $\mu$ can determine the position of the 2-D Gaussian distribution and the covariance matrix $\Sigma$ can influence the shape of the 2-D Gaussian distribution:
 \begin{equation}
    \mu = \begin{bmatrix}\mu_{X}\\ \mu_{Y}\end{bmatrix}, \Sigma = \begin{bmatrix}\sigma_{X}^2 & \rho\sigma_{X}\sigma_{Y}\\ \rho\sigma_{X}\sigma_{Y} & \sigma_{Y}^2\end{bmatrix},
 \label{MGD}
 \end{equation}
 \noindent where $\rho$ is the correlation coefficient between $X$ and $Y$, $\mu_{X}$ represents the $X$ position of the 2-D Gaussian that equals to $X$ position of the GPS data $x_{gps}$,  $\mu_{Y}$ represents the $Y$ position of the 2-D Gaussian that equals to the $Y$ position of the GPS data $y_{gps}$, $\sigma_{X}^2$ represents the extension of 2-D Gaussian along $X$ axis and $\sigma_{Y}^2$ represents the extension of 2-D Gaussian along $Y$ axis \cite{prince2012computer}.

 As a result, the relationships between $\sigma_{X}^2$, $\sigma_{Y}^2$ and $R$ can be described as below:
 \begin{equation}
  \sigma_{X}^2 = \sigma_{Y}^2 = R^{2},
 \label{MGD_1}
 \end{equation}
 \noindent Further, the relationships between $\sigma_{X}^2$, $\sigma_{Y}^2$ and AIS data that describes the dimension specification of the target USV can be transformed as below:
 \begin{equation}
  \sigma_{X}^2 = \sigma_{Y}^2 = [\frac{\zeta \cdot(A + B + C + D)}{2}]^{2}.
 \label{MGD_2}
 \end{equation}

 \begin{figure}[t!]
 \centering
 \includegraphics[width=1 \linewidth]{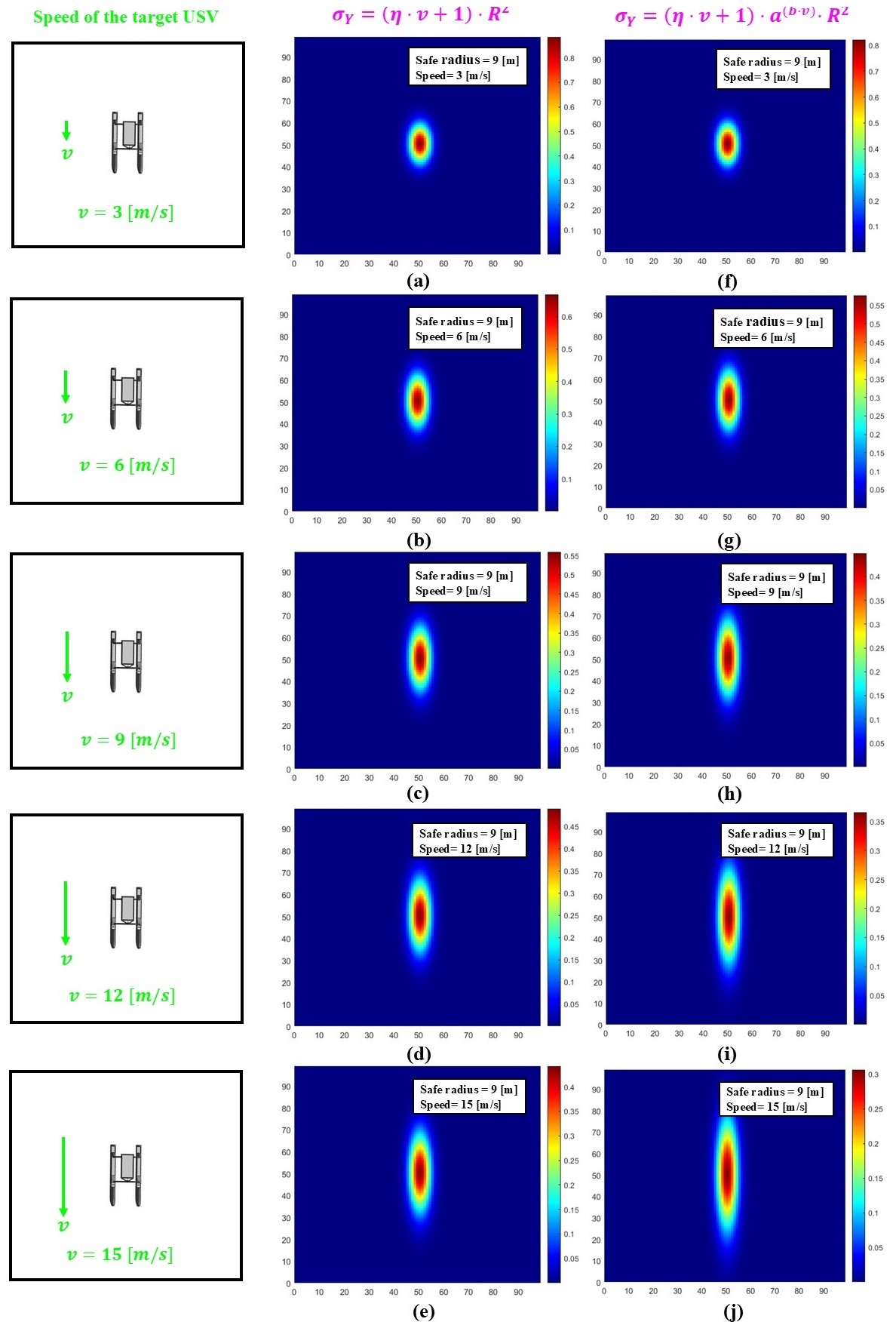}
 \centering
 \caption{A comparison of the models constructed by (\ref{vessel_speed}) and (\ref{vessel_speed_with_calibration}). The sub-figures (a) - (e) demonstrate the models constructed by (\ref{vessel_speed}). The sub-figures (f) - (j) demonstrate the models constructed by (\ref{vessel_speed_with_calibration}). In both comparative groups, the speed of the target USV changes from 0 [m/s] to 15 [m/s]. In the colour bars, a larger value indicates a higher risk and a smaller value indicates a lower risk.}
 \label{calibration_performance} 
 \end{figure}

 \begin{figure}[t!]
 \centering
 \includegraphics[width=1 \linewidth]{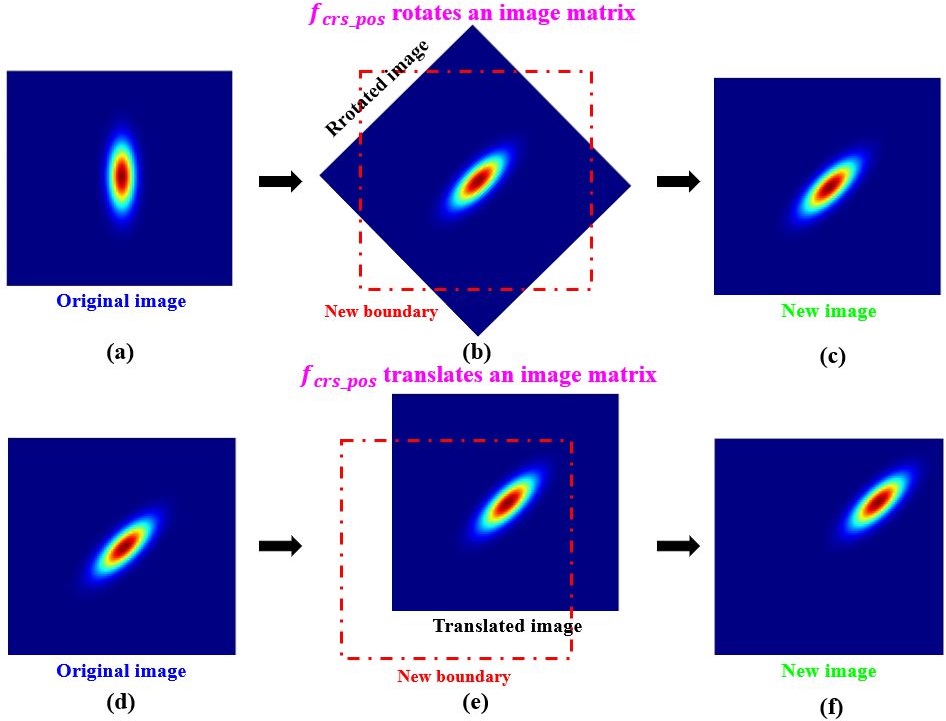}
 \centering
 \caption{(a) - (c) demonstrates of the working process of $f_{crs\_pos}$ rotates an image matrix. (d) - (f) demonstrates of the working process of $f_{crs\_pos}$ translates an image matrix. The risk at the point close to the dark red region is relatively high; on the other hand, the risk at the point close to the dark blue region is relatively low.}
 \label{image_rotation_translation} 
 \end{figure}
 
 When constructing the safe area of a USV, the USV is always placed at the centre of the image and will be translated to the correct position according to GPS data later. To provide a more intuitive understanding of the proposed dynamic obstacle area modelling method, the examples in Fig. \ref{2D_Gaussian_different_areas} demonstrates the modelling results of the safe areas of USVs with different dimensions. Meanwhile, Table \ref{Table:vessels} provides the dimension specifications of the modelled USVs.
  
\subsection{Dynamic obstacle velocity modelling}
\label{Dynamic Obstacle Velocity Modelling}

 To reduce the risk of colliding with a target USV, it is necessary to integrate the position and velocity information of the target USV into the model constructed in Section \ref{Dynamic Obstacle Area Modelling}. To achieve this goal, we need to use the target USV position from the GPS data and target USV speed and course from the AIS data. By incorporating the information into a new model, the proposed D-GPMP2 algorithm can use the new model to generate a trajectory to avoid any potential movement of the target USV.  
 
 The speed of a target USV from the AIS data can be integrated into the dynamic obstacle area model by using the following equation:
 \begin{equation}
    \begin{bmatrix}\sigma_{X}^2\\ \sigma_{Y}^2\end{bmatrix} = \begin{bmatrix}R^{2}\\ (\eta \cdot v+1) \cdot R^{2}\end{bmatrix},
 \label{vessel_speed}
 \end{equation}
 \noindent where $\eta$ is a adjustable positive scaling factor and $v$ is the target USV speed from the AIS data. Now the safe area constructed by the method described in Section \ref{Dynamic Obstacle Area Modelling} is extended along the positive half axis of $Y$ axis as a result of the influence of (\ref{vessel_speed}). Nevertheless, we notice that the increased length in each meter per second of the model constructed by (\ref{vessel_speed}) gradually decreases with the speed gradually increasing. Therefore, we add a self-defined exponential function into (\ref{vessel_speed}) to calibrate this error:
 \begin{equation}
    \begin{bmatrix}\sigma_{X}^2\\ \sigma_{Y}^2\end{bmatrix} = \begin{bmatrix} R^{2}\\ (\eta \cdot v+1) \cdot a^{(b\cdot v)} \cdot R^{2}\end{bmatrix}
 \label{vessel_speed_with_calibration}
 \end{equation}

 \begin{figure}[t!]
 \centering
 \includegraphics[width=1 \linewidth]{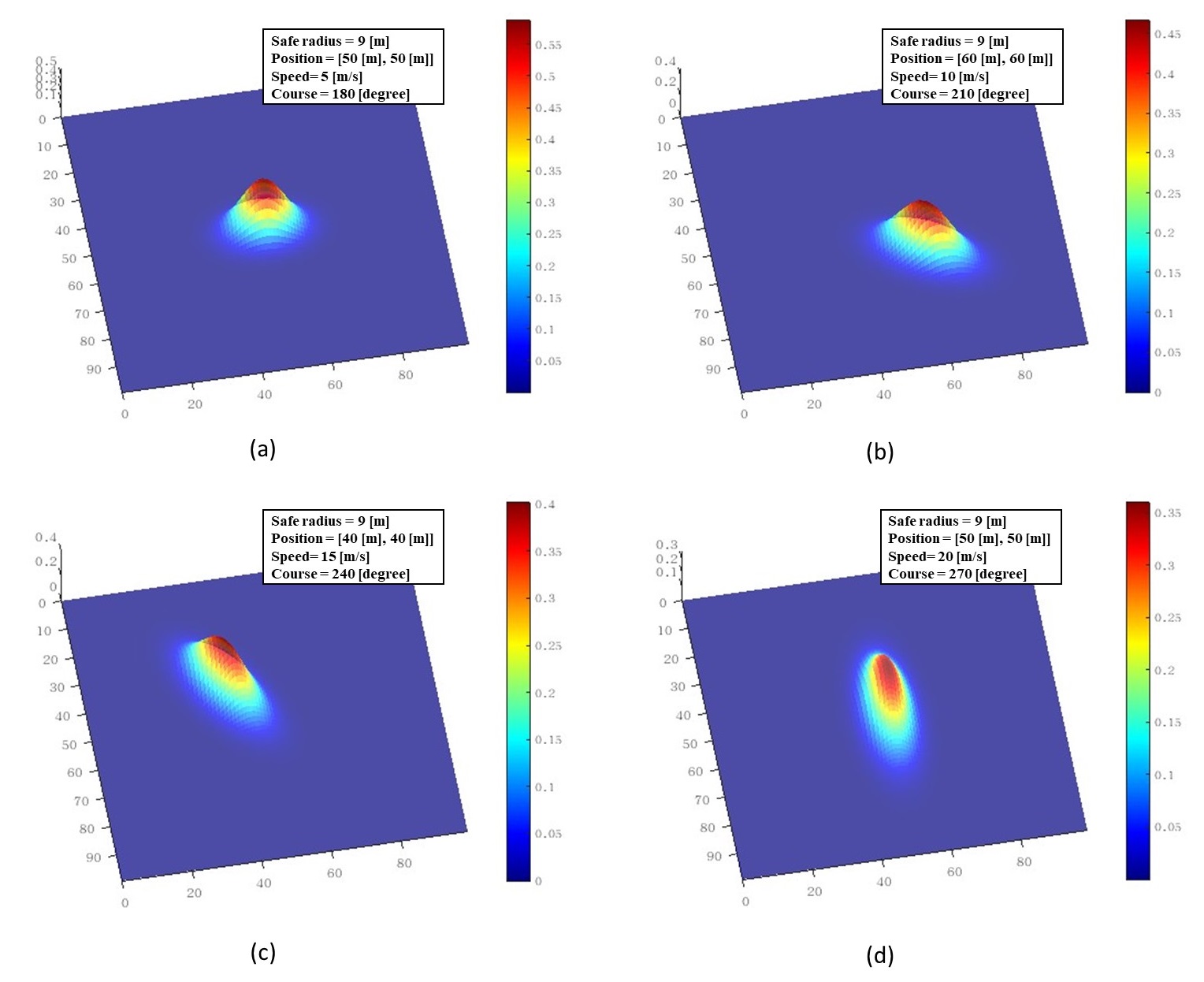}
 \centering
 \caption{Demonstration of the modelling results of target USVs with different velocities and positions in MATLAB from the 2-D perspective and the 3-D perspective. In the colour bars, a larger value indicates a higher risk and a smaller value indicates a lower risk.}
 \label{2D_Gaussian_different_velocities} 
 \end{figure} 

 As demonstrated in the first comparative group in Fig. \ref{calibration_performance} (a) - (e), the models constructed by (\ref{vessel_speed}) no longer change significantly with the speed increases. Nevertheless, as demonstrated in the second comparative group in Fig. \ref{calibration_performance} (f) - (j), the models constructed by (\ref{vessel_speed_with_calibration}) with the self-defined exponential function reduce the error. It is worth noting that the model constructed by our method is extended both the forward and backward directions of velocity. Due to the possibility of small errors in GPS positioning, the model expands in the backward direction of velocity to ensure the obstacle can be avoided from behind safely.
 
 Next, the model constructed by (\ref{vessel_speed_with_calibration}) needs to be rotated towards the target USV course and translated to the target USV position. To achieve these goals, a function is designed as below:
 \begin{equation}
  M'_{N \times N} = f_{crs\_pos}(M_{N \times N}, \alpha, x_{gps}, y_{gps}),
  \label{function}
 \end{equation}
 \noindent where $M_{N \times N}$ is a $N$-by-$N$ image matrix that contains the model constructed by (\ref{vessel_speed_with_calibration}). More specifically, there are two main steps when the function in (\ref{function}) rotates an image matrix: 1) rotating the original image around its centre with a specified angle $\alpha$, where $\alpha$ equals to the difference between the target USV course and the positive half axis of $Y$ axis and 2) interpolating zero values into the pixels that are inside the new boundary of the image matrix but outside the rotated image. Similarly, there are two main steps when the function designed in (\ref{function}) translates an image matrix: 1) translating the centre of the original image to the target USV position [$x_{gps}$, $y_{gps}$] and 2) interpolating zero values into the pixels that are inside the new boundary of the image matrix but outside the translated image.

 Fig. \ref{image_rotation_translation} provides a series of examples to give a more intuitive understanding of the working process of $f_{crs\_pos}$, while Fig. \ref{2D_Gaussian_different_velocities} demonstrates the modelling results of target USVs with different velocities and positions based on (\ref{vessel_speed_with_calibration}) and (\ref{function}). The detailed process of constructing dynamic obstacle model is provided in Algorithm \ref{alg: dynamic obstacle modelling}.

 \begin{algorithm}[t!]
 \SetAlgoLined
 \textbf{Input:} Length of the target USV ($l$), width of the target USV ($w$), speed of the target USV ($v$), position of the target USV [$x_{gps}$, $y_{gps}$] and the difference between the target USV course and the positive half axis of $Y$ axis ($\alpha$) \\
 Compute the safe radius of the target USV by using equation (\ref{safe_radius}); \\
 Perform dynamic area modelling for the target USV by using equation (\ref{MGD}) and (\ref{MGD_1}); \\
 Perform dynamic velocity modelling for the target USV by using equation (\ref{vessel_speed}); \\
 Calibrate the constructed dynamic velocity model by using equation (\ref{vessel_speed_with_calibration}); \\
 Adjust the position and orientation of the calibrated dynamic obstacle model by using equation (\ref{function}); \\
 \textbf{Output:} Dynamic obstacle likelihood l($\theta$; \textit{$e_{dy\_obs}$})\\
 \caption{\hl{Dynamic obstacle modelling}}
 \label{alg: dynamic obstacle modelling}
 \end{algorithm}

\subsection{A comparison between multivariate Gaussian and artificial potential field}
\begin{figure}[t!]
 \centering
 \includegraphics[width=1 \linewidth]{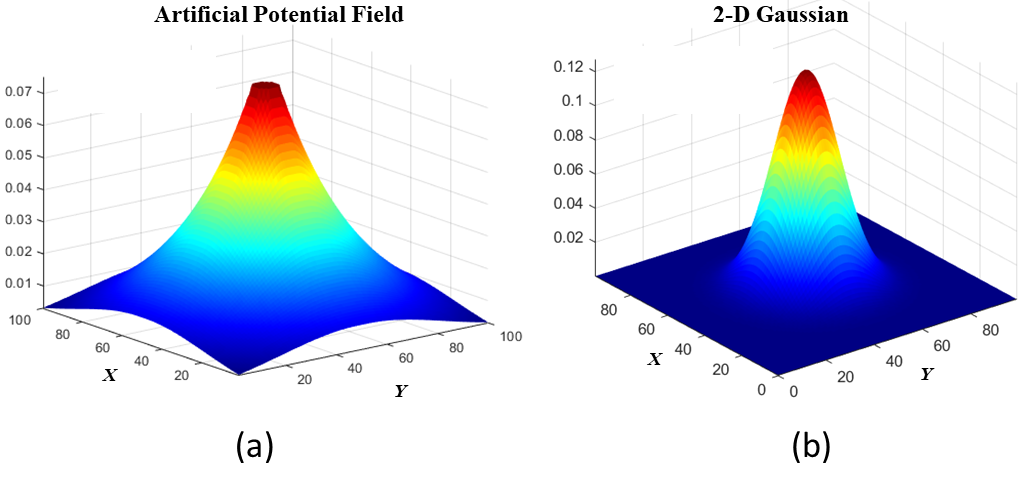}
 \centering
 \caption{(a) Object modelled by the artificial potential field. (b) Object modelled by the 2-D Gaussian. In both sub-figures, a larger value indicates a higher risk and a smaller value indicates a lower risk.}
 \label{comparison_with_APF} 
 \end{figure}

 It was found that the artificial potential field algorithm has been used as a mainstream method of building object models in many previous studies deal with dynamic object avoidance in motion planning and a famous example can be found in \cite{falanga2020dynamic}. In order to prove the effectiveness of the proposed method and a comparison of an object modelled by using the artificial potential field and 2-D Gaussian is provided in Fig. \ref{comparison_with_APF}. It is worth noting that the 2-D Gaussian can generate a model with a clearer boundary and a clear boundary helps to generate a trajectory that can avoid dynamic objects in a more efficient manner.

 \subsection{Dynamic obstacle likelihood} 

 Based on the aforementioned information, a dynamic obstacle likelihood that considers 1) the safe areas, 2) velocities and 3) positions of the target USVs can be written as a distribution in the exponential family as \cite{meng2022anisotropic, mukadam2018continuous}:
 \begin{equation}
    l(\theta; \textit{$e_{dy\_obs}$}) = exp\{ -\frac{1}{2}||h_{dy\_obs}(\theta)||^{2}_{\sigma_{dy\_obs}}\},
 \end{equation}
 \noindent where $e_{dy\_obs}$ represents the event of avoiding dynamic obstacles, the definition of matrix $\sigma_{\textit{dy\_obs}}$ is similar with $\sigma_{\textit{st\_obs}}$ and $h_{dy\_obs}(\theta)$ is a vector-valued cost function:
 \begin{equation}
         h_{dy\_obs}(\theta_{i}) = [f_{crs\_pos}(f_{2}(\theta_{i}),\alpha, x_{gps}, y_{gps}))],
 \label{dynamic_obstacle_likelihood}
 \end{equation}
 \noindent where the mean vector $\mu$ and the covariance matrix $\Sigma$ of the 2-D Gaussian probability density function $f_{2}$ are:
 \begin{equation}
 \footnotesize
    \mu = \begin{bmatrix}x_{gps}\\ y_{gps}\end{bmatrix}, 
 \end{equation}
 \begin{equation}
 \footnotesize
    \Sigma = \begin{bmatrix}\frac{1}{4} \zeta^2 (A + B + C + D)^{2} & 0\\ 0 & \frac{1}{4} \zeta^2 (\eta v+1) a^{bv} (A + B + C + D)^{2}\end{bmatrix}.
 \end{equation}

 \subsection{MAP estimation}
 Based on the static obstacle likelihood and the dynamic obstacle likelihood, the Bayesian inference of the proposed D-GPMP2 algorithm can then be defined as:   
 \begin{equation}
 \begin{gathered}
 \theta_{0:N}^{*} = \mathop{\arg\max}_{\theta} \; p(\theta)p(e_{st\_obs}|\theta)p(e_{dy\_obs}|\theta).
 \end{gathered}
 \label{MAP2}
 \end{equation}

 In the proposed D-GPMP2 algorithm, we also use the Levenberg–Marquardt method to solve the least squares problem \cite{meng2022anisotropic, mukadam2018continuous, levenberg1944method}.

 \section{D-GPMP2-COLREGs}
 \label{D-GPMP2-COLREGs}
 \hl{This section extends the proposed D-GPMP2 algorithm through COLREG rules integration and proposes a variant named D-GPMP2-COLREGs algorithm.
 
 The COLREG rules define behaviours for USVs in overtaking, head-on and crossing scenarios} \cite{colreg}:

 \begin{itemize}
     \item \hl{Overtaking: the overtaking USV is required to keep out of the way of the USV that is being overtaken.}
     \item \hl{Head-on: if two USVs are meeting on reciprocal or nearly reciprocal courses, each USV is required to adjust its course to starboard and passes on the port side of the other.} 
     \item \hl{Crossing: if two USVs are crossing, the USV that has the other on its own starboard side is required to keep out of the way. This USV is also required to avoid crossing ahead of the other USV.}
 \end{itemize}

 \hl{D-GPMP2-COLREGs algorithm is proposed to enable the proposed D-GPMP2 algorithm to generate trajectories that comply with COLREGs rules. The pseudocode of D-GPMP2-COLREGs is detailed in Algorithm} \ref{alg:D-GPMP2-COLREGs}: \hl{colregs\_region() is a function that calculates feasible regions for trajectory planning, ensuring compliance with COLREGs rules. random\_angle() is a function that generates a uniformly distributed random angle within given bounds. factor\_graph() constructs a factor graph based on a series of acquired data. levenberg\_marquardt() is a function that performs gradient-based optimisation via the Levenberg Marquardt method. The feasible regions for D-GPMP2-COLREGs algorithm to generate trajectories in overtaking, head-on, and crossing scenarios in colregs\_region() are detailed below}:

 \begin{algorithm}[t!]
 \SetAlgoLined
 \textbf{Input:} Start state $\theta_{0}$, goal state $\theta_{N}$ and dynamic obstacle position [$x_{gps}$, $y_{gps}$] \\
 \textbf{Precompute} static obstacle likelihood l($\theta$; \textit{$e_{st\_obs}$}) and dynamic obstacle likelihood l($\theta$; \textit{$e_{dy\_obs}$}) \\
 S = colregs\_region($\theta_{0}$, [$x_{gps}$, $y_{gps}$]);\\
 $\psi$  = random\_angle($\theta_{0}$, S);\\
 $G$ = factor\_graph($\theta_{0}$, $\theta_{N}$, $\psi$, l($\theta$; \textit{$e_{st\_obs}$}), l($\theta$; \textit{$e_{dy\_obs}$}));\\
 $\theta^{*}$ = levenberg\_marquardt($G$);\\
 \textbf{Output:} Optimal path $\theta^{*}$\\
 \caption{\hl{D-GPMP2-COLREGs (Dynamic Gaussian process motion planner 2 based on COLREGs rules)}}
 \label{alg:D-GPMP2-COLREGs}
 \end{algorithm}

 \begin{figure}[t!]
 \centering
 \includegraphics[width=1 \linewidth]{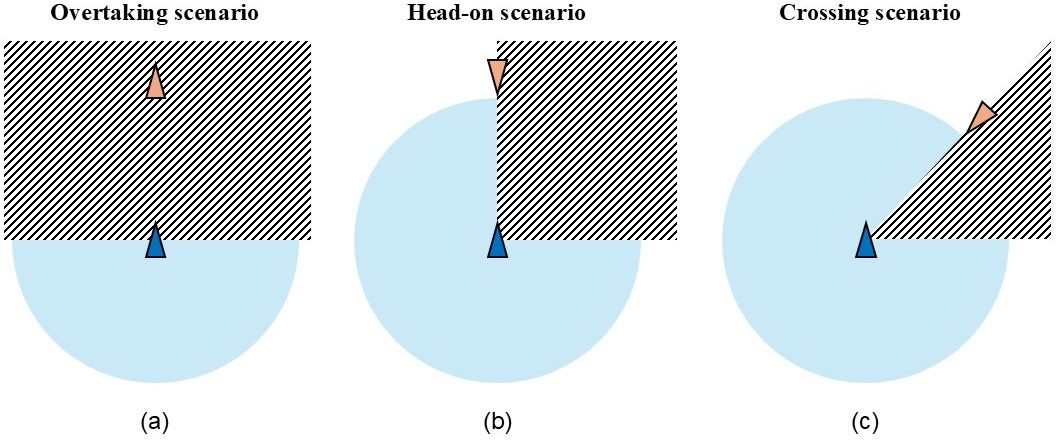}
 \centering
 \caption{\hl{The regions that D-GPMP2-COLREGs algorithm generates trajectories in overtaking, head-on and crossing scenarios. (a) overtaking scenario, (b) head-on scenario and (c) crossing scenario. The blue triangle indicates our USV, while the orange triangle represents the encountered USV. The shaded regions represents the regions that D-GPMP2-COLREGs algorithm can generate trajectories.}}
 \label{shadow_region} 
 \end{figure}

 \begin{itemize}
     \item \hl{Overtaking scenario: trajectories can be generated on either sides of the encountered USV as demonstrated in Fig.} \ref{shadow_region}(a).
     \item \hl{Head-on scenario: trajectories can be generated on the starboard side of our USV as demonstrated in Fig.} \ref{shadow_region}(b).
     \item \hl{Crossing scenario: trajectories can be generated on the starboard side of our USV and the port side of the encountered USV as demonstrated in Fig.} \ref{shadow_region}(c).
 \end{itemize}

 \hl{Since the coordinate systems in the real world and in the high-fidelity virtual world in Gazebo differ, it is required to calibrate the error between these coordinate systems before calculating any angles, as we have done in} \cite{meng2022fully}. 

 \begin{figure*}[t!]
 \centering
 \includegraphics[width=1 \linewidth]{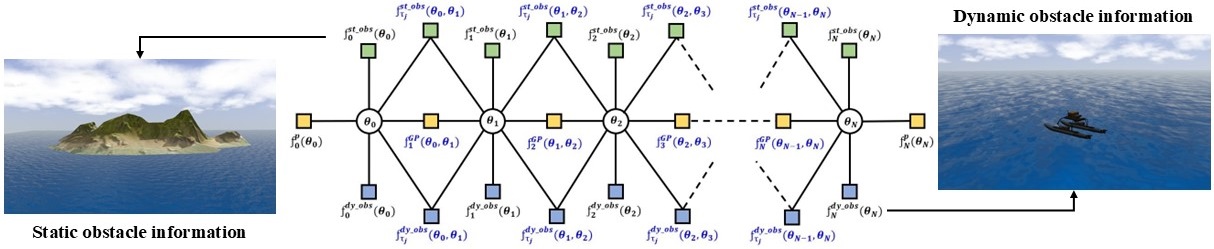}
 \centering
 \caption{Overall structure of the factor graph. $f^{p}_{0}(\theta_{0})$ and $f^{p}_{N}(\theta_{N})$ represent the regular prior factors, $f^{gp}_{i}(\theta_{i}, \theta_{i+1})$ represent the GP prior factors, $f^{st\_obs}_{i}(\theta_{i})$ represent the static obstacle likelihood factors, $f^{st\_obs}_{\tau_{j}}(\theta_{i}, \theta_{i+1})$ represent the interpolated static obstacle factors, $f^{dy\_obs}_{i}(\theta_{i})$ represent the dynamic obstacle likelihood factors and $f^{dy\_obs}_{\tau_{j}}(\theta_{i}, \theta_{i+1})$ represent the interpolated dynamic obstacle factors. \hl{$f^{st\_obs}_{i}(\theta_{i})$ and $f^{st\_obs}_{\tau_{j}}(\theta_{i}, \theta_{i+1})$ contains static obstacle information from the surrounding environment. $f^{dy\_obs}_{i}(\theta_{i})$ and $f^{dy\_obs}_{\tau_{j}}(\theta_{i}, \theta_{i+1})$ contains dynamic obstacle information from the surrounding environment.}}
 \label{factor_graph} 
 \end{figure*}

\section{Factor graph}
\label{Factor graph}
 In the proposed D-GPMP2 and D-GPMP2-COLREGs algorithms, factor graph is used as a optimisation tool to deal with the MAP estimation and a new structure of it is designed to calculate the optimised trajectory to avoid static and dynamic obstacles simultaneously. More specifically, this new structure includes three categories of factors \cite{meng2022anisotropic, mukadam2018continuous}:
 \begin{itemize}
     \item Prior factors such as 1) the regular prior factors [$f^{p}_{0}(\theta_{0})$ and $f^{p}_{N}(\theta_{N})$] and 2) the GP prior factors [$f^{gp}_{i}(\theta_{i}, \theta_{i+1})$].
     \item Static obstacle likelihood factors such as 1) the regular static obstacle likelihood factors [$f^{st\_obs}_{i}(\theta_{i})$] and 2) the interpolated static obstacle likelihood factors [$f^{st\_obs}_{\tau_{j}}(\theta_{i}, \theta_{i+1})$].
     \item Dynamic obstacle likelihood factors such as 1) the regular dynamic obstacle likelihood factors [$f^{dy\_obs}_{i}(\theta_{i})$] and 2) the interpolated dynamic obstacle likelihood factors [$f^{dy\_obs}_{\tau_{j}}(\theta_{i}, \theta_{i+1})$].
 \end{itemize}
 
 Fig. \ref{factor_graph} demonstrates the overall structure of the factor graph used in this article. Generally, a factor graph $G$ is a bipartite graph and can be defined as:
  \begin{equation}
   G = \{\Theta, \mathcal{F}, \mathcal{E}\},
  \end{equation}
 \noindent where $\Theta$ are a set of variable nodes, $\mathcal{F}$ are factor nodes attached to the corresponding variable nodes via $\mathcal{E}$ that are edges that connect the variable nodes and factor nodes. The factorisation of the posterior in our problem can be formulated as:
  \begin{equation}
    p(\theta|e_{s}, e_{d}) \propto \prod^{M}_{m=1}f_{m}(\Theta_{m}),
  \end{equation}
 \noindent where $f_{m}$ are factors on variable subset $\Theta_{m}$ \cite{meng2022anisotropic, mukadam2018continuous}.

 There are static obstacle likelihood factors in our motion planning problem regardless of whether there are static obstacles. And if there are no static obstacles, a signed distance field is generated based on an empty binary map.

\section{Simulations and discussions}
\label{Simulations and discussions}
 This section demonstrates the performance of the proposed motion planning algorithms on dynamic obstacle avoidance under different situations \hl{in a series of qualitative and quantitative simulations}.

 \subsection{Simulation details}
 In general, the proposed motion planning algorithms have been tested using \hl{seven} different categories of benchmark simulations: 
 
 \begin{itemize}
     \item The first benchmark simulation validated the performance of the proposed D-GPMP2 algorithm when different weights of dynamic obstacle likelihood were chosen.
     \item The second benchmark simulation validated the performance of the proposed D-GPMP2 algorithm when the velocity of the dynamic obstacle was adjusted. More specifically, two quantitative sub-simulations were conducted with only the speed or direction of the dynamic obstacle adjusted.
     \item The third benchmark simulation validated the performance of the proposed D-GPMP2 algorithm when the area of the dynamic obstacle was adjusted.
     \item The fourth benchmark simulation validated the performance of the proposed D-GPMP2 algorithm in different multiple dynamic obstacles situations.
     \item The fifth benchmark simulation validated the performance of the proposed D-GPMP2 algorithm when both static and dynamic obstacles required to be avoided in complex maritime environments.
     \item \hl{The sixth benchmark simulation compared the performance of the proposed D-GPMP2 algorithm with an existing method (DWA) in complex maritime environments.}
     \item \hl{The seventh benchmark simulation demonstrated the performance of the proposed D-GPMP2-COLREGs algorithm in overtaking, head-on and crossing scenarios.}
 \end{itemize}

 \begin{table}[t!]
 \caption{Specification of the used hardware platform.}
 \label{Table:2}
 \centering
 \footnotesize
 \begin{tabular}{c c c}
 \hline
 \textbf{Device} & \textbf{Description} & \textbf{Quantity} \\ \hline
 {Processor} & \makecell{2.3-GHz Intel Core i7-11800H} & 16 \\ 
 {RAM} & 8 GB & 1 \\ \hline
 \end{tabular}
 \end{table}
      
\begin{figure}[t!]
 \centering
 \includegraphics[width=1 \linewidth]{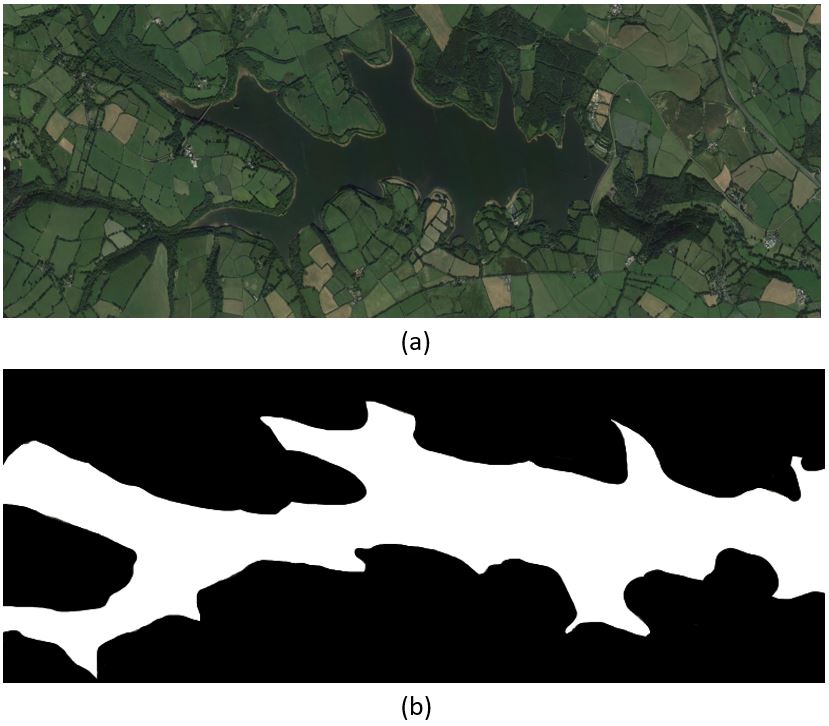}
 \centering
 \caption{(a) demonstrates the geographical map of Roadford Lake, Devon, UK \cite{googleearth}. (b) demonstrates the binary map about the Lake generated by us. A square part of the binary map in (b) is used in the fifth benchmark simulation.}
 \label{lake_map} 
 \end{figure}

 In all the benchmark simulations, the proposed motion planning algorithms were initialised with a CV-SL trajectory based on the system dynamics model described in Section \ref{System dynamics model}. Table \ref{Table:2} demonstrates the specification of the used hardware platform for running the benchmark simulations, while Fig. \ref{lake_map} demonstrates the maritime map used in the fifth benchmark simulation. Meanwhile, the obstacle angle in the following benchmark simulations is the target USV course. \hl{The parameters used in the following simulations were identified based on empirical measurements.}

 \subsection{System dynamics model}
 \label{System dynamics model}
The dynamics of the robot platform is represented with the double integrator linear system with additional white noise on acceleration \cite{meng2022anisotropic, dong2016motion, mukadam2018continuous}. A CV prior is used because it can minimise acceleration along the trajectory, reduce energy consumption and increase the smoothness of the trajectory. Thus, the trajectory is generated by LTV-SED with parameters:
 
 \begin{equation}
    A = \begin{bmatrix}0 & I\\ 0 & 0\end{bmatrix}, x(t) = \begin{bmatrix}r(t)\\v(t)\end{bmatrix}, F(t) = \begin{bmatrix}0\\I\end{bmatrix}, u(t) = 0, 
 \end{equation}
 and given $\Delta t_{i} = t_{i+1} - t_{i}$,
  \begin{equation}
  \small
    \Phi(t,s) = \begin{bmatrix}I & (t-s)I\\ 0 & I\end{bmatrix}, Q_{i,i+1} = \begin{bmatrix} \frac{1}{3}\Delta t_{i}^{3}Q_{C} & \frac{1}{2}\Delta t_{i}^{2}Q_{C} \\ \frac{1}{2}\Delta t_{i}^{2}Q_{C} & \Delta t_{i}Q_{C} \end{bmatrix},
 \end{equation}
 The sample trajectories from the GP prior is centred around a zero-acceleration trajectory.

 \subsection{Simulation with different weights of dynamic obstacle likelihood}
 In this subsection, we demonstrate the influence of the weight of the dynamic obstacle likelihood in the proposed D-GPMP2 algorithm. 
 
\begin{figure}[t!]
 \centering
 \includegraphics[width=1 \linewidth]{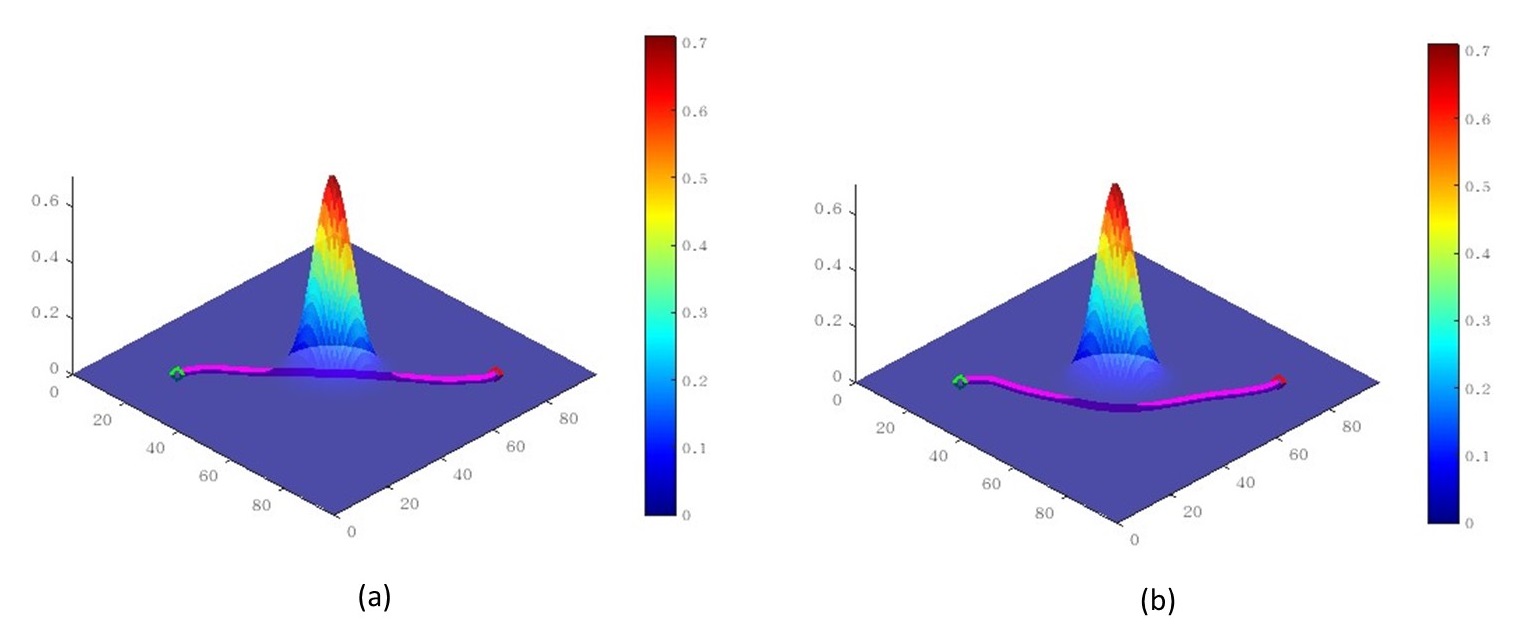}
 \centering
 \caption{Demonstration of the simulation results with different weights of dynamic obstacle likelihood: From (a) to (b), the weight of dynamic obstacle likelihood is 0.5 and 0.005, respectively. The start position is represented in green, the goal position is represented in red and the trajectory is represented in purple. In the colour bars, a larger value indicates a higher risk and a smaller value indicates a lower risk.}
 \label{first_simulation} 
 \end{figure}

 We utilised the proposed D-GPMP2 algorithm to solve a motion planning problem with a dynamic obstacle located at a point between the start position and goal position and moving towards the start position. The major parameters used in the proposed D-GPMP2 algorithm when solving this motion planning problem is demonstrated in Table \ref{Table:3}. In this simulation, we gradually adjusted the weight of the dynamic obstacle likelihood from a relatively larger value to a relatively smaller value and observed the change of the generated trajectory. As can be seen in Fig. \ref{first_simulation} (a) - (b), the capability of the proposed D-GPMP2 algorithm to avoid dynamic obstacle gradually improved when the weight of the dynamic obstacle likelihood gradually decreased.

 To summarise, the performance of the proposed D-GPMP2 algorithm to avoid dynamic obstacle is adjustable. Further, this characteristic can help us choose different obstacle avoidance strategies in different situations. An example would be:
 \begin{itemize}
     \item We are able to enhance the capability of the proposed D-GPMP2 algorithm of avoiding dynamic obstacle when the obstacle is moving towards our platform with a relatively fast speed.
     \item We are also able to improve the travel efficiency by maintaining the generated path straight if the dynamic obstacle is actively avoiding our platform.
 \end{itemize}

 \begin{table}[t!]
 \caption{List of parameters for the simulation with different weights of dynamic obstacle likelihood.}
 \label{Table:3}
 \centering
 \footnotesize
 \begin{tabular}{c c c}
 \hline
 \textbf{Number} & \textbf{Parameter name} & \textbf{Parameter value} \\ \hline
  1 & Map size & 100 $\times$ 100 [m$^2$] \\ 
  2 & Start position & [20, 20] \\ 
  3 & Goal position & [80, 80] \\ 
  4 & Obstacle position & [50, 50] \\
  5 & obstacle speed & 2 [m/s] \\ 
  6 & Obstacle angle & 225 [degree] \\ 
  7 & Obstacle length & 6 [m] \\ 
  8 & Obstacle width & 3 [m] \\ 
  9 & Safe radius & 9 [m] \\ \hline
 \end{tabular}
 \end{table}

 \subsection{Simulation with a dynamic obstacle with different velocities}
 In this subsection, we demonstrate the performance of the proposed D-GPMP2 algorithm when adjusting the velocity of the dynamic obstacle and maintaining the other parameters constant. More specifically, two quantitative simulations were conducted in order to demonstrate the influence of the speed and direction of the velocity of the dynamic obstacle on the generated trajectory.

 \begin{figure}[t!]
 \centering
 \includegraphics[width=1 \linewidth]{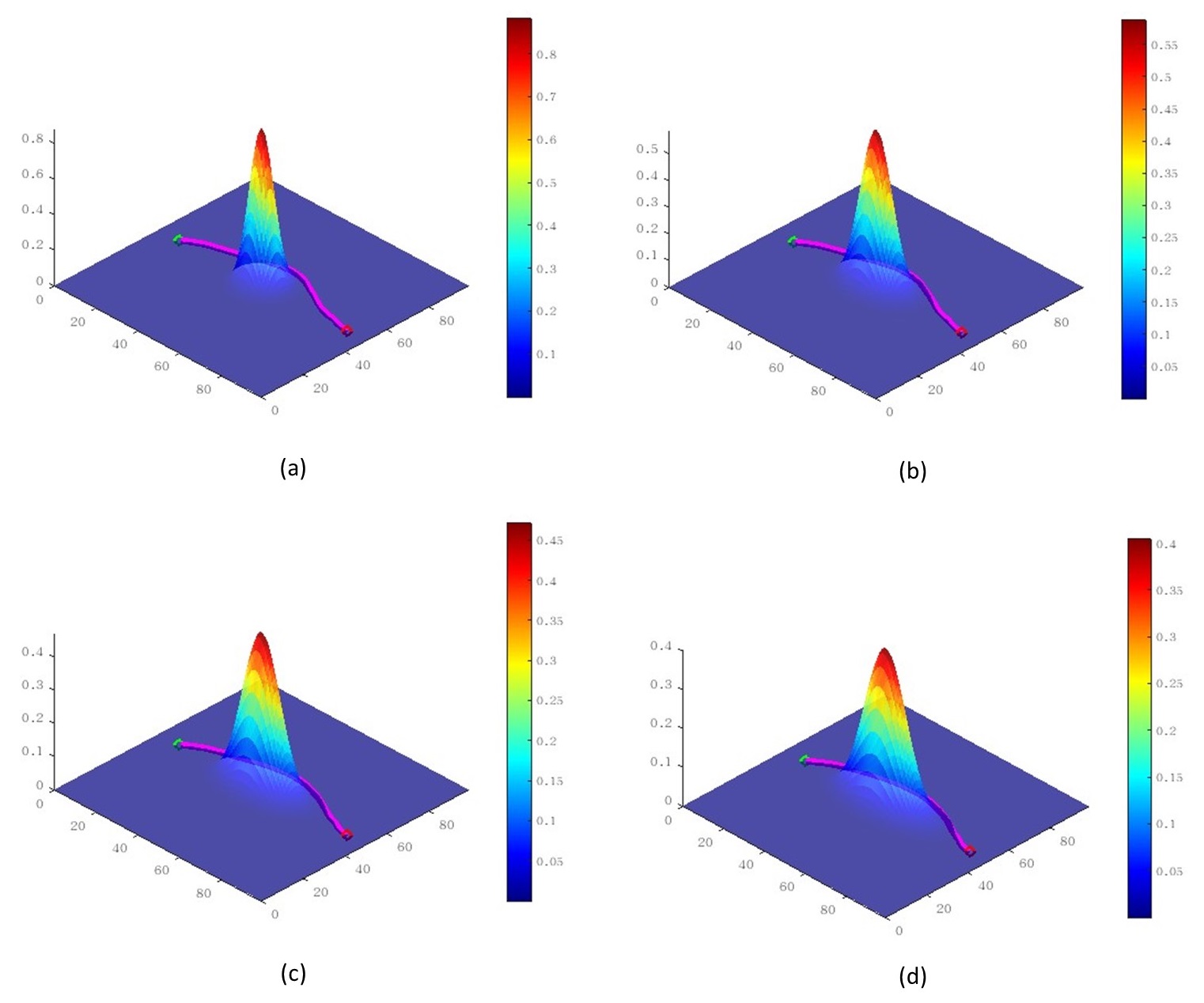}
 \centering
 \caption{Demonstration of the simulation results with a dynamic obstacle with different speeds: From (a) to (d), the velocity of the dynamic obstacle is 0 [m/s], 5 [m/s], 10 [m/s] and 15 [m/s], respectively. The start position is represented in green, the goal position is represented in red and the trajectory is represented in purple. In the colour bars, a larger value indicates a higher risk and a smaller value indicates a lower risk.}
 \label{second_simulation_1} 
 \end{figure}

 \begin{table}[t!]
 \caption{List of parameters for the simulation with a dynamic obstacle with different speeds.}
 \label{Table:4}
 \centering
 \footnotesize
 \begin{tabular}{ c c c }
 \hline
 \textbf{Number} & \textbf{Parameter name} & \textbf{Parameter value} \\ \hline
  1 & Map size & 100 $\times$ 100 [m$^2$] \\ 
  2 & Start position & [10, 50] \\ 
  3 & Goal position & [90, 50] \\ 
  4 & Obstacle position & [50, 50] \\ 
  5 & Obstacle angle & 270 [degree] \\ 
  6 & Obstacle length & 6 [m] \\ 
  7 & Obstacle width & 3 [m] \\ 
  8 & Safe radius & 9 [m] \\ 
  9 & \makecell{Dynamic obstacle\\likelihood weight} & 0.005 \\ \hline
 \end{tabular}
 \end{table}
 
 \subsubsection{Simulation with a dynamic obstacle with different speeds}
 This simulation used the proposed D-GPMP2 algorithm to solve a motion planning problem with a dynamic obstacle located between the start position and goal position and moving towards the start position. Table \ref{Table:4} demonstrates the major parameters used in the proposed D-GPMP2 algorithm when solving this motion planning problem. While running different simulation trials, we gradually increased the speed of the dynamic obstacle from a relatively smaller value to a relatively larger value and observed the change on the generated trajectory. As can be seen in the simulation results from Fig. \ref{second_simulation_1} (a) - (d), increasing the speed of the dynamic obstacle caused the generated trajectory to avoid the dynamic obstacle earlier.

 In summary, the proposed D-GPMP2 algorithm can automatically find a suitable moment to avoid dynamic obstacle according to the speed of the dynamic obstacle. This characteristic ensures the safety of the generated trajectory as well as improves the efficiency of the proposed D-GPMP2 algorithm. As a result, the proposed D-GPMP2 algorithm is adaptable for motion planning problems involving dynamic obstacles with various speeds.

 \begin{figure}[t!]
 \centering
 \includegraphics[width=1 \linewidth]{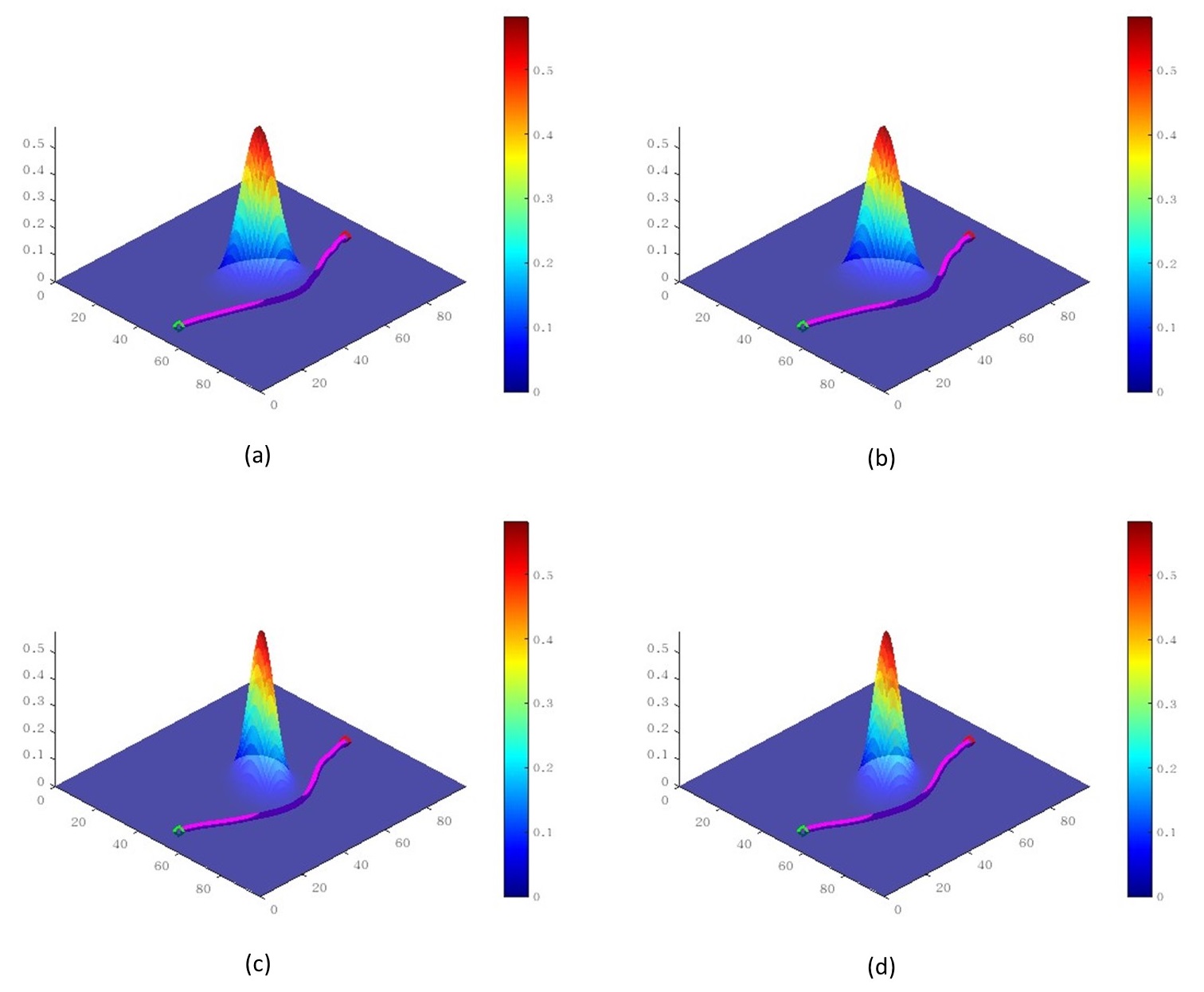}
 \centering
 \caption{Demonstration of the simulation results with a dynamic obstacle with different directions: From (a) to (d), the angle of the dynamic obstacle is 210 [degree], 240 [degree], 300 [degree] and 330 [degree], respectively. The start position is represented in green, the goal position is represented in red and the trajectory is represented in purple. In the colour bars, a larger value indicates a higher risk and a smaller value indicates a lower risk.}
 \label{second_simulation_3} 
 \end{figure}

 \begin{table}[t!]
 \caption{List of parameters for the simulation with dynamic obstacles with different directions.}
 \label{Table:5}
 \centering
 \footnotesize
 \begin{tabular}{ c c c }
 \hline
 \textbf{Number} & \textbf{Parameter name} & \textbf{Parameter value} \\ \hline
  1 & Map size & 100 $\times$ 100 [m$^2$] \\ 
  2 & Start position & [50, 10] \\ 
  3 & Goal position & [50, 90] \\ 
  4 & Obstacle position & [50, 50] \\ 
  5 & Obstacle speed & 5 [m/s] \\ 
  6 & Obstacle length & 6 [m] \\ 
  7 & Obstacle width & 3 [m] \\ 
  8 & Safe radius & 9 [m] \\ 
  9 & \makecell{Dynamic obstacle\\likelihood weight} & 0.005 \\ \hline
 \end{tabular}
 \end{table}

 \subsubsection{Simulation with a dynamic obstacle with different directions}
 The proposed D-GPMP2 algorithm was used in this simulation to solve a motion planning problem with a dynamic obstacle passing by between the start position and goal position with different angles. The major parameters used in the proposed D-GPMP2 algorithm when solving this motion planning problem are demonstrated in Table \ref{Table:5}. In this simulation, we gradually adjusted the angle of the dynamic obstacle when it was passing by between the start position and goal position and observed the change on the generated trajectory. As can be seen in the simulation results in Fig. \ref{second_simulation_3} (a) - (d), the curved part of the generated trajectory is always located at a position in the opposite direction of the velocity of the dynamic obstacle.

 In summary, the proposed D-GPMP2 algorithm can generate a trajectory to avoid a dynamic obstacle through bypassing it from behind at a suitable moment. This characteristic reduces the collision risk and improves the efficiency of the proposed D-GPMP2 algorithm. Consequently, the proposed motion planning algorithm is adaptable to problems involving dynamic obstacles moving in different directions.

 \begin{figure}[t!]
 \centering
 \includegraphics[width=1 \linewidth]{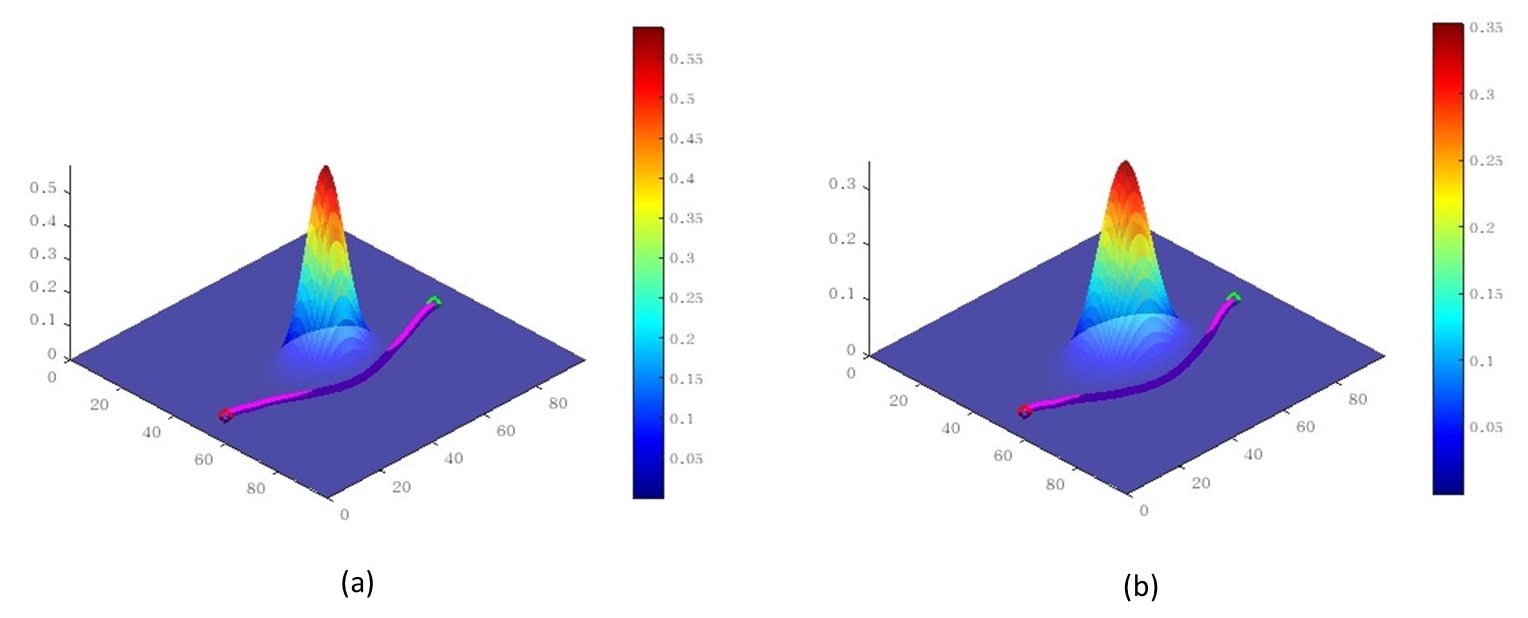}
 \centering
 \caption{Demonstration of the simulation results with dynamic obstacles with different areas: (a) The length of the dynamic obstacle is 6 [m], the width of the dynamic obstacle is 3 [m] and the safe radius of the dynamic obstacle is 9 [m]. (b) The length of the dynamic obstacle is 9 [m], the width of the dynamic obstacle is 6 [m] and the safe radius of the dynamic obstacle is 15 [m]. The start position is represented in green, the goal position is represented in red and the trajectory is represented in purple. In the colour bars, a larger value indicates a higher risk and a smaller value indicates a lower risk.}
 \label{second_simulation_2} 
 \end{figure}

 \begin{table}[t!]
 \caption{List of parameters for the simulation with dynamic obstacles with different areas.}
 \label{Table:6}
 \centering
 \footnotesize
 \begin{tabular}{ c c c }
 \hline
 \textbf{Number} & \textbf{Parameter name} & \textbf{Parameter value} \\ \hline
  1 & Map size & 100 $\times$ 100 [m$^2$] \\ 
  2 & Start position & [50, 90] \\ 
  3 & Goal position & [50, 10] \\ 
  4 & Obstacle position & [50, 50] \\ 
  5 & Obstacle angle & 180 [degree] \\ 
  6 & Obstacle speed & 5 [m/s] \\ 
  7 & \makecell{Dynamic obstacle\\likelihood weight} & 0.005 \\ \hline
 \end{tabular}
 \end{table}

 \begin{figure}[t!]
 \centering
 \includegraphics[width=1 \linewidth]{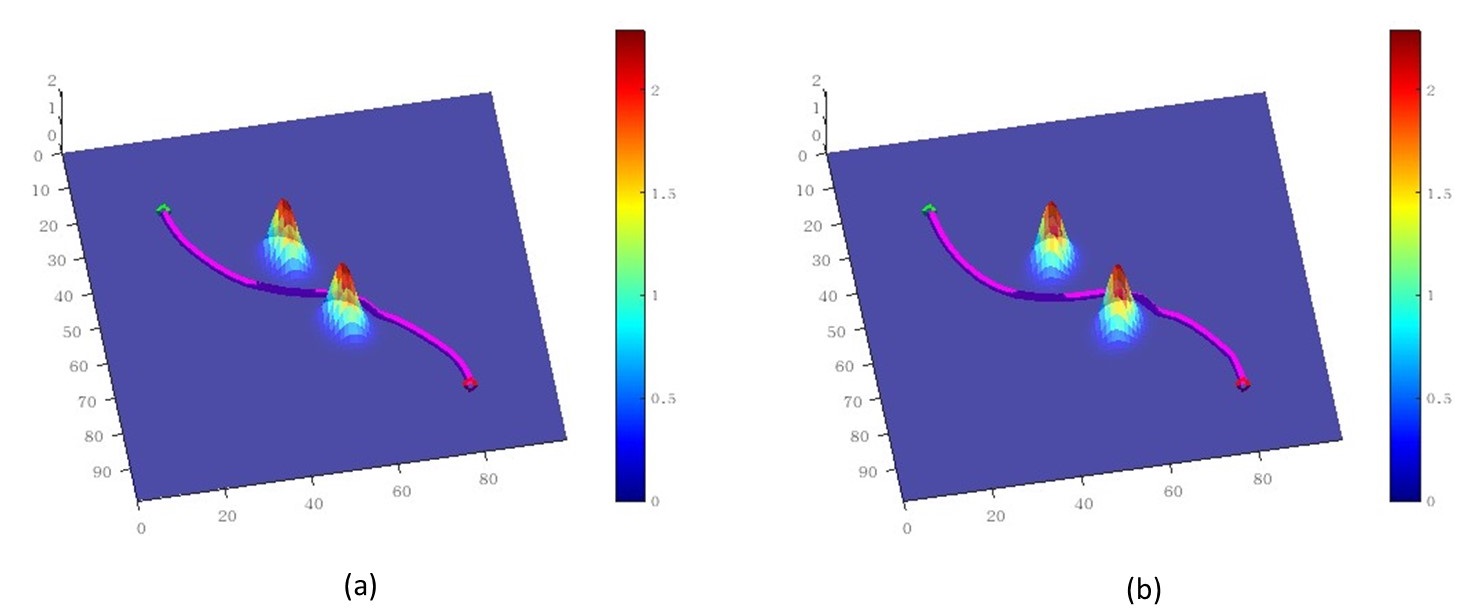}
 \centering
 \caption{Demonstration of the simulation results with multiple dynamic obstacles with different areas and velocities: (a) The position of the first dynamic obstacle is [40, 45] and the position of the second dynamic obstacle is [60, 55]. Furthermore, the angle of the two dynamic obstacles is 225 [degree]. (b) The position of the first dynamic obstacle is [40, 45] and the position of the second dynamic obstacle is [60, 55]. Furthermore, the angle of the two dynamic obstacles is 135 [degree] and 315 [degree], respectively. The start position is represented in green, the goal position is represented in red and the trajectory is represented in purple. In the colour bars, a larger value indicates a higher risk and a smaller value indicates a lower risk.}
 \label{third_simulation} 
 \end{figure}

 \begin{table}[t!]
 \caption{List of parameters for the simulation with multiple dynamic obstacles.}
 \label{Table:7}
 \centering
 \footnotesize
 \begin{tabular}{ c c c }
 \hline
 \textbf{Number} & \textbf{Parameter name} & \textbf{Parameter value} \\ \hline
  1 & Map size & 100 $\times$ 100 [m$^2$] \\ 
  2 & Start position & [20, 20] \\ 
  3 & Goal position & [80, 80] \\ 
  4 & Obstacle speed & 3 [m/s] \\ 
  5 & Obstacle length & 1.5 [m] \\ 
  6 & Obstacle width & 1 [m] \\ 
  7 & Safe radius & 2.5 [m] \\ 
  8 & \makecell{Dynamic obstacle\\likelihood weight} & 0.005 \\ \hline
 \end{tabular}
 \end{table}
 
 \subsection{Simulation with dynamic obstacles with different areas}
 In this subsection, we investigate how the proposed D-GPMP2 algorithm performs if the dynamic obstacle area changes while the other parameters remain constant. 
 
 In this simulation, the proposed D-GPMP2 algorithm was applied to a motion planning problem with a dynamic obstacle between the start position and goal position and moving towards the start position. Table \ref{Table:6} demonstrates the major parameters used in the proposed D-GPMP2 algorithm when solving this motion planning problem. We observed the change on the generated trajectory after gradually increasing the area of the dynamic obstacle from a relatively smaller value to a relatively larger value during a series of simulation trials. The simulation results in Fig. \ref{second_simulation_2} (a) - (b) demonstrate that an increase in the area of the dynamic obstacle results in an earlier avoidance of the obstacle on the generated trajectory. The reason for this is that a dynamic obstacle with a similar speed and a larger area is more likely to collide. 

 In summary, the D-GPMP2 algorithm can guarantee the safety and efficiency of obstacle avoidance by taking into account the area of dynamic obstacles. 

 \subsection{Simulation with multiple dynamic obstacles}
 In this subsection, we demonstrate the performance of the proposed D-GPMP2 algorithm in handling multiple dynamic obstacles.
 
 In this simulation, two motion planning problems involving two dynamic obstacles were resolved using the proposed D-GPMP2 algorithm:
 \begin{itemize}
     \item In the first motion planning problem, two dynamic obstacles move towards the start position continuously.
     \item In the second motion planning problem, two dynamic obstacles pass between the start position and goal position in opposite directions.
 \end{itemize}  
 
 Table \ref{Table:7} demonstrates the major parameters used in the proposed D-GPMP2 algorithm in these motion planning problems. The simulation results in Fig. \ref{second_simulation_1} (a) - (b) demonstrate that the generated trajectories can successfully avoid the dynamic obstacles in both the motion planning problems.

 In summary, the proposed D-GPMP2 algorithm is adaptable not only to single dynamic obstacle situations, but also to multiple dynamic obstacle situations.

 \subsection{Simulation with static and dynamic obstacles.}
 This subsection describes the performance of the proposed D-GPMP2 algorithm when dealing with static and dynamic obstacles together in a complex and dynamic maritime environment. 

\begin{figure}[t!]
 \centering
 \includegraphics[width=1 \linewidth]{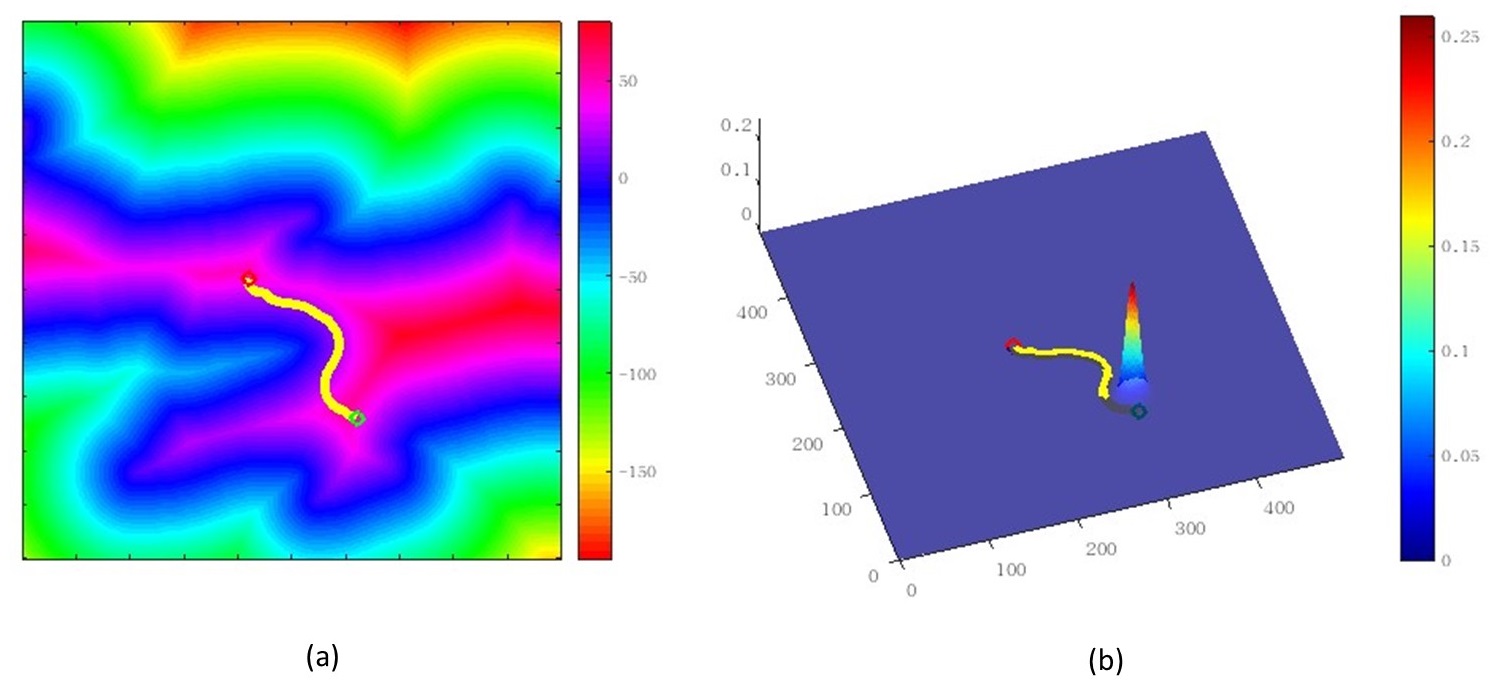}
 \centering
 \caption{Demonstration of the simulation result with both static and dynamic obstacles. (a) The trajectory avoids the static obstacle. (b) The trajectory avoids the dynamic obstacle. The start position is represented in green, the goal position is represented in red and the trajectory is represented in yellow. In the colour bar of sub-figure (a), a larger value indicates a lower risk and a smaller value indicates a higher risk. In the colour bar of sub-figure (b), a larger value indicates a higher risk and a smaller value indicates a lower risk.}
 \label{fourth_simulation} 
 \end{figure}

 This simulation used the proposed D-GPMP2 algorithm to solve a motion planning problem that involves dynamic obstacles passing between the start position and goal position as well as static obstacles such as islands surrounding the area. Table \ref{Table:8} demonstrates the major parameters used in the proposed D-GPMP2 algorithm when solving this motion planning problem. The simulation result in Fig. \ref{fourth_simulation} demonstrates that the generated trajectory can successfully avoid both the static and dynamic obstacles. Before reaching the goal position, the generated trajectory first avoids the dynamic obstacle, which is relatively close to the start position, then navigates around the raised part of the island. Furthermore, it is worth noting that the proposed motion planning algorithm can generate a safe, smooth and relatively short path within a very short period of time according to the summary of the simulation result provided in Table \ref{Table:9}. Specifically, the total execution time of the main steps in the used motion planning algorithm in this simulation is less than 50 [ms] in 500 $\times$ 500 [m$^2$] map.

 In summary, the proposed D-GPMP2 algorithm can handle static and dynamic obstacles simultaneously. This characteristic significantly improves the efficiency of the proposed D-GPMP2 algorithm and broadens the potential application scenarios. 

 \begin{table}[t!]
 \caption{List of parameters for the simulation with static and dynamic obstacles.}
 \label{Table:8}
 \centering
 \footnotesize
 \begin{tabular}{ c c c }
 \hline
 \textbf{Number} & \textbf{Parameter name} & \textbf{Parameter value} \\ \hline
  1 & Map size & 500 $\times$ 500 [m$^2$] \\ 
  2 & Start position & [310, 130] \\ 
  3 & Goal position & [210, 260] \\ 
  4 & Obstacle position & [310, 153] \\ 
  5 & Obstacle speed & 3 [m/s] \\ 
  6 & Obstacle length & 15 [m] \\ 
  7 & Obstacle width & 8 [m] \\ 
  8 & Safe radius & 23 [m] \\ 
  9 & Obstacle angle & 225 [degree] \\ 
  10 & \makecell{Dynamic obstacle\\likelihood weight} & 0.003 \\ 
  11 & \makecell{Static obstacle\\likelihood weight} & 0.05 \\ 
  12 & \makecell{Safe distance} & 20 [m] \\ \hline
 \end{tabular}
 \end{table}

 \begin{table}[t!]
 \caption{Summary of the simulation result with static and dynamic obstacles}
 \label{Table:9}
 \centering
 \footnotesize
 \begin{tabular}{ c c c }
 \hline
 \textbf{Number} & \textbf{Measurement} & \textbf{Value} \\ \hline
  1 & Path length & 201.0 [m] \\ 
  2 & \makecell{Static obstacle likelihood\\time cost} & 3.9 [ms] \\ 
  3 & \makecell{Dynamic obstacle likelihood\\time cost} & 9.7 [ms] \\ 
  4 & Factor graph time cost & 13.9 [ms] \\ 
  5 & MAP estimation time cost & 11.8 [ms] \\ \hline
 \end{tabular}
 \end{table}

 \subsection{Benchmark comparison with DWA}

 \begin{figure}[t!]
 \centering
 \includegraphics[width=1 \linewidth]{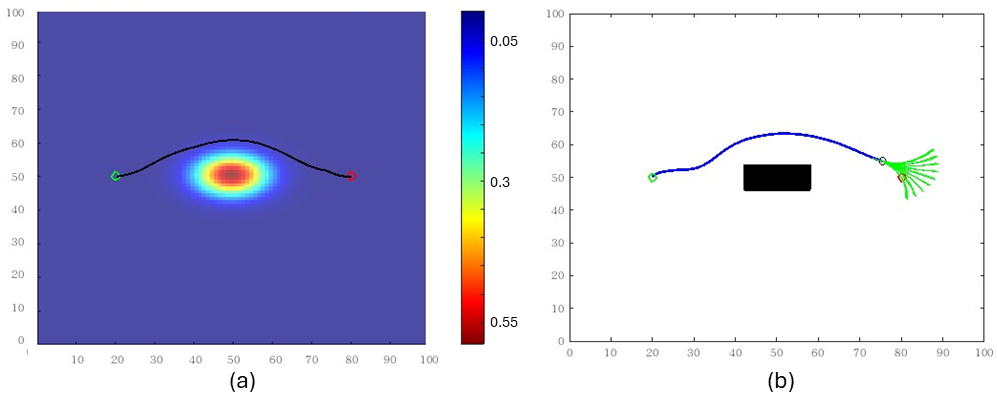}
 \centering
 \caption[Demonstration of the simulation result of the benchmark comparison with DWA.]{\hl{Demonstration of the simulation result of the benchmark comparison with DWA. (a) The trajectory generated by the proposed D-GPMP2 algorithm. (b) The trajectory generated by DWA. The start positions are represented as green circles, the goal positions are represented as red circles, the trajectory generated by the proposed D-GPMP2 algorithm is represented as black line, the trajectory generated by DWA is represented as blue line and the potential local trajectories of DWA are represented as green arrows. In the colour bar of sub-figure (a), a larger value indicates a higher risk and a smaller value indicates a lower risk.}}
 \label{benchmark_comparison} 
 \end{figure}

 \begin{table}[t!]
 \caption{\hl{List of parameters of the proposed D-GPMP2 algorithm in benchmark comparison.}}
 \label{Table:quantitative_dgpmp2 chapter 6}
 \centering
 \footnotesize
 \begin{tabular}{ c  c  c }
 \hline
 \textbf{Number} & \textbf{Parameter name} & \textbf{Parameter value} \\ \hline
  1 & Map size & 100 $\times$ 100 [m$^2$] \\ 
  2 & Start position & [20, 50] \\ 
  3 & Goal position & [80, 50] \\ 
  4 & Obstacle position & [50, 50] \\
  5 & obstacle speed & 5 [m/s] \\ 
  6 & Obstacle angle & 270 [degree] \\ 
  7 & Obstacle length & 6 [m] \\ 
  8 & Obstacle width & 3 [m] \\ 
  9 & Safe radius & 9 [m] \\
  10 & Likelihood weight & 0.01 \\
  \hline
 \end{tabular}
 \end{table}

\begin{table}[t!]
 \caption{\hl{List of parameters of DWA in benchmark comparison.}}
 \label{Table:quantitative_dwa chapter 6}
 \centering
 \footnotesize
 \begin{tabular}{ c  c  c }
 \hline
 \textbf{Number} & \textbf{Parameter name} & \textbf{Parameter value} \\ \hline
  1 & Map size & 100 $\times$ 100 [m$^2$] \\ 
  2 & Start position & [20, 50] \\ 
  3 & Goal position & [80, 50] \\ 
  4 & Obstacle position & [50, 50] \\
  5 & Safe radius & 5 [m] \\
  6 & Sample interval & 0.5 [s] \\
  7 & Prediction time & 3 [s] \\
  8 & Linear velocity resolution & 0.05 [m/s] \\
  9 & Angular velocity resolution & 0.1 [rad/s] \\
  \hline
 \end{tabular}
 \end{table}

\begin{table}[t!]
 \caption{\hl{Summary of the simulation result in the benchmark comparison between the proposed D-GPMP2 and DWA.}}
 \label{Table:benchmark_result}
 \centering
 \footnotesize
 \begin{tabular}{  c  c  c  c  }
 \hline
 \textbf{Number} & \textbf{Measurement} & D-GPMP2 & DWA \\ \hline
  1 & Path length & \textbf{64.4} [m]  &  67.8 [m] \\ 
  2 & Time cost & \textbf{28.9} [ms] & 68.2 [ms] \\
 \hline
 \end{tabular}
 \end{table}

 \hl{This subsection compares the performance of the proposed D-GPMP2 algorithm with an existing obstacle avoid strategy (DWA) in an environment with a dynamic obstacle.}
 
 \hl{Only a quantitative comparison with DWA is presented, as mainstream motion planning algorithms like A* and RRT* cannot meet real-time requirements for dynamic obstacle avoidance in many cases due to their computational inefficiency. Comprehensive benchmarking data highlighting the performance differences on computational speed between GP-based algorithms and mainstream motion planning algorithms are detailed in} \cite{meng2022anisotropic, meng2022fully}.

 \hl{This benchmark simulation compared the performance between the proposed D-GPMP2 algorithm and DWA in an obstacle avoidance scenario, where a moving obstacle exists between the start and goal positions.} Table \ref{Table:quantitative_dgpmp2 chapter 6} \hl{demonstrates the parameters used in the proposed D-GPMP2 algorithm, while} Table \ref{Table:quantitative_dwa chapter 6} \hl{demonstrates the parameters used in the DWA.} \hl{To align with the dynamic obstacle modelling in D-GPMP2, the obstacle model in DWA was expanded along its movement direction.} Fig. \ref{benchmark_comparison} \hl{demonstrates the simulation results of the proposed D-GPMP2 and DWA, while} Table \ref{Table:benchmark_result} \hl{compares both the generated path length and total time required for generating the path to avoid the moving obstacle.} \hl{It is worth noting that D-GPMP2 can generate a shorter path with reduced computation time for dynamic obstacle avoidance compared with DWA.}

 \hl{In summary, the quantitative comparison with a mainstream obstacle avoidance strategy (DWA) demonstrates D-GPMP2's superior performance in dynamic obstacle avoidance.}

 \subsection{COLREGs-compliant benchmark simulation}

 \hl{This subsection provides a quantitative performance analysis of the proposed D-GPMP2-COLREGs algorithm for COLREGs-constrained dynamic obstacle avoidance in maritime environments.}

\begin{figure}[t!]
 \centering
 \includegraphics[width=1 \linewidth]{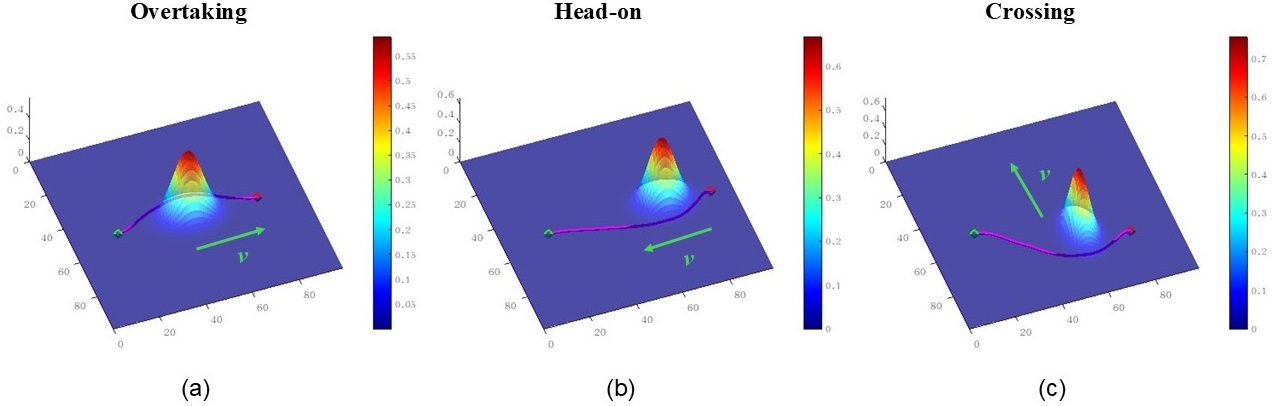}
 \centering
 \caption{\hl{Demonstration of the results of COLREGs-compliant benchmark simulation. (a) Overtaking result, (b) Head-on result and (c) Crossing result. In all sub-figures, our USV is positioned at the starting point and moving along the positive Y-axis direction. The green arrows indicate the moving directions of the dynamic obstacles.}}
 \label{d-gpmp2_colregs_simulation} 
 \end{figure}

 The benchmark demonstrates that the D-GPMP2-COLREGs algorithm effectively handles dynamic obstacle avoidance in maritime scenarios while strictly complying with COLREGs regulations. Fig. \ref{d-gpmp2_colregs_simulation} (a) demonstrates the result of the proposed D-GPMP2-COLREGs algorithm in an overtaking simulation scenario. Because COLREGs don't strictly specify the overtaking side, our USV overtakes the encountered USV by passing on its starboard side. Fig. \ref{d-gpmp2_colregs_simulation} (b) demonstrates the result of the proposed D-GPMP2-COLREGs algorithm in a head-on simulation scenario. In compliance with the COLREGs rule for head-on situations, which requires port-side passing, the proposed D-GPMP2-COLREGs algorithm generates a regulation-compliant trajectory. Fig. \ref{d-gpmp2_colregs_simulation} (c) demonstrates the result of the proposed D-GPMP2-COLREGs algorithm in a crossing simulation scenario. As can be seen in this sub-figure, our USV passes behind the encountered vessel's port side in compliance with COLREGs rules.

 \hl{In summary, the proposed D-GPMP2-COLREGs algorithm can effectively avoid dynamic obstacles while following COLREGs rules.}
 
 \begin{table}[t!]
 \caption{\hl{List of parameters of the proposed D-GPMP2-COLREGs algorithm in COLREGs-compliant benchmark simulation (overtaking).}}
 \label{Table:overtaking}
 \centering
 \footnotesize
 \begin{tabular}{ c  c  c }
 \hline
 \textbf{Number} & \textbf{Parameter name} & \textbf{Parameter value} \\ \hline
  1 & Map size & 100 $\times$ 100 [m$^2$] \\ 
  2 & Start position & [50, 20] \\ 
  3 & Goal position & [50, 80] \\ 
  4 & Obstacle position & [50, 50] \\
  5 & obstacle speed & 5 [m/s] \\ 
  6 & Obstacle angle & 0 [degree] \\ 
  7 & Obstacle length & 6 [m] \\ 
  8 & Obstacle width & 3 [m] \\ 
  9 & Safe radius & 9 [m] \\
  10 & Likelihood weight & 0.01 \\
  \hline
 \end{tabular}
 \end{table}

 \begin{table}[t!]
 \caption{\hl{List of parameters of the proposed D-GPMP2-COLREGs algorithm in COLREGs-compliant benchmark simulation (head-on).}}
 \label{Table:head-on}
 \centering
 \footnotesize
 \begin{tabular}{ c  c  c }
 \hline
 \textbf{Number} & \textbf{Parameter name} & \textbf{Parameter value} \\ \hline
  1 & Map size & 100 $\times$ 100 [m$^2$] \\ 
  2 & Start position & [50, 20] \\ 
  3 & Goal position & [50, 90] \\ 
  4 & Obstacle position & [50, 70] \\
  5 & obstacle speed & 3 [m/s] \\ 
  6 & Obstacle angle & 180 [degree] \\ 
  7 & Obstacle length & 6 [m] \\ 
  8 & Obstacle width & 3 [m] \\ 
  9 & Safe radius & 9 [m] \\
  10 & Likelihood weight & 0.01 \\
  \hline
 \end{tabular}
 \end{table}

  \begin{table}[t!]
 \caption{\hl{List of parameters of the proposed D-GPMP2-COLREGs algorithm in COLREGs-compliant benchmark simulation (crossing).}}
 \label{Table:crossing}
 \centering
 \footnotesize
 \begin{tabular}{ c  c  c }
 \hline
 \textbf{Number} & \textbf{Parameter name} & \textbf{Parameter value} \\ \hline
  1 & Map size & 100 $\times$ 100 [m$^2$] \\ 
  2 & Start position & [50, 20] \\ 
  3 & Goal position & [70, 80] \\ 
  4 & Obstacle position & [65, 60] \\
  5 & obstacle speed & 5 [m/s] \\ 
  6 & Obstacle angle & 270 [degree] \\ 
  7 & Obstacle length & 5 [m] \\ 
  8 & Obstacle width & 2 [m] \\ 
  9 & Safe radius & 7 [m] \\
  10 & Likelihood weight & 0.01 \\
  \hline
 \end{tabular}
 \end{table}

 \section{Implementation in ROS}
 \label{Implementation in ROS}

 This section demonstrates the practical performance of the proposed D-GPMP2 algorithm in a dynamic obstacle avoidance mission in a high-fidelity maritime environment in ROS. 

\subsection{Implementation details}
\begin{figure}[t!]
 \centering
 \includegraphics[width=1 \linewidth]{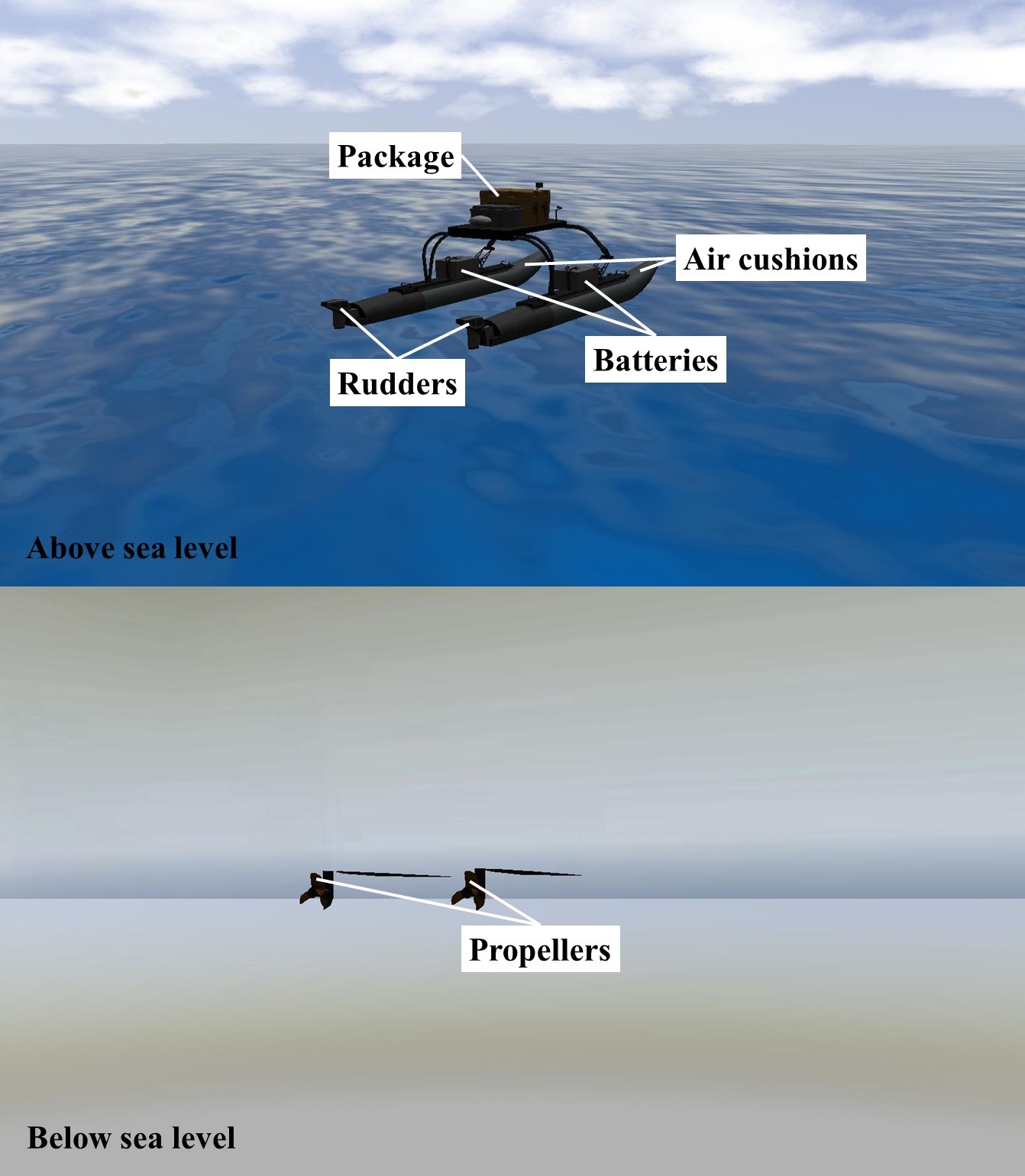}
 \centering
 \caption{Demonstration of the selected USV platform in the high-fidelity maritime environment in ROS Gazebo \cite{bingham2019toward}.}
 \label{wam-v-model} 
 \end{figure}

\begin{figure}[t!]
 \centering
 \includegraphics[width=1 \linewidth]{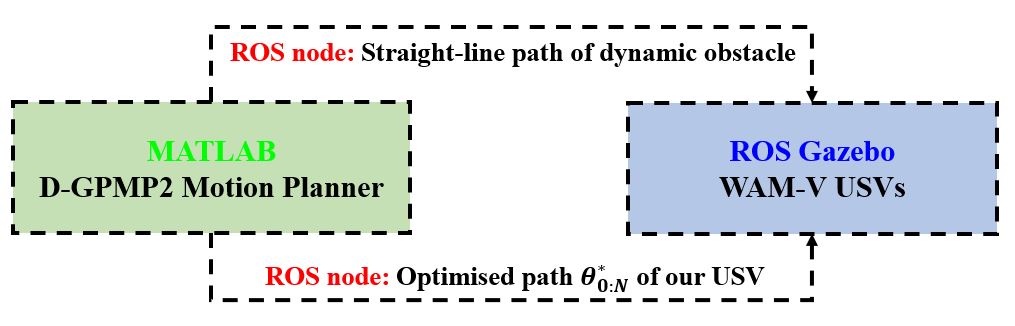}
 \centering
 \caption{The connection between MATLAB (the proposed D-GPMP2 algorithm) and ROS Gazebo (WAM-V 20 USVs).}
 \label{matlab_ros} 
 \end{figure}

 Compared with practical experiments in real world, the implementation in ROS has a lower cost and is more efficient. \hl{The implementation in ROS considers the physical properties such as gravity, buoyancy, hydrodynamic forces, inertia, deformation caused by collision as well as other necessary factors to achieve exceptional correlation with actual experimental results.} A widely-used catamaran, the WAM-V 20 USV, is provided by the high-fidelity maritime environment and the environment is provided by \cite{bingham2019toward}. Fig. \ref{wam-v-model} provides an intuitive overview of the WAM-V 20 USV above and below sea level in the ROS Gazebo virtual world. \hl{An additional WAM-V 20 USV was implemented in the simulation environment for the following dynamic obstacle avoidance mission.} As demonstrated in Fig. \ref{matlab_ros}, the proposed D-GPMP2 algorithm was run on MATLAB and connected with the WAM-V 20 USVs in ROS Gazebo through ROS node. \hl{In real-world scenarios and high-fidelity simulation environments, dynamic obstacle avoidance through path following requires non-negligible time cost and cannot be performed instantaneously.} 

 \begin{figure}[t!]
 \centering
 \includegraphics[width=1.0\linewidth]{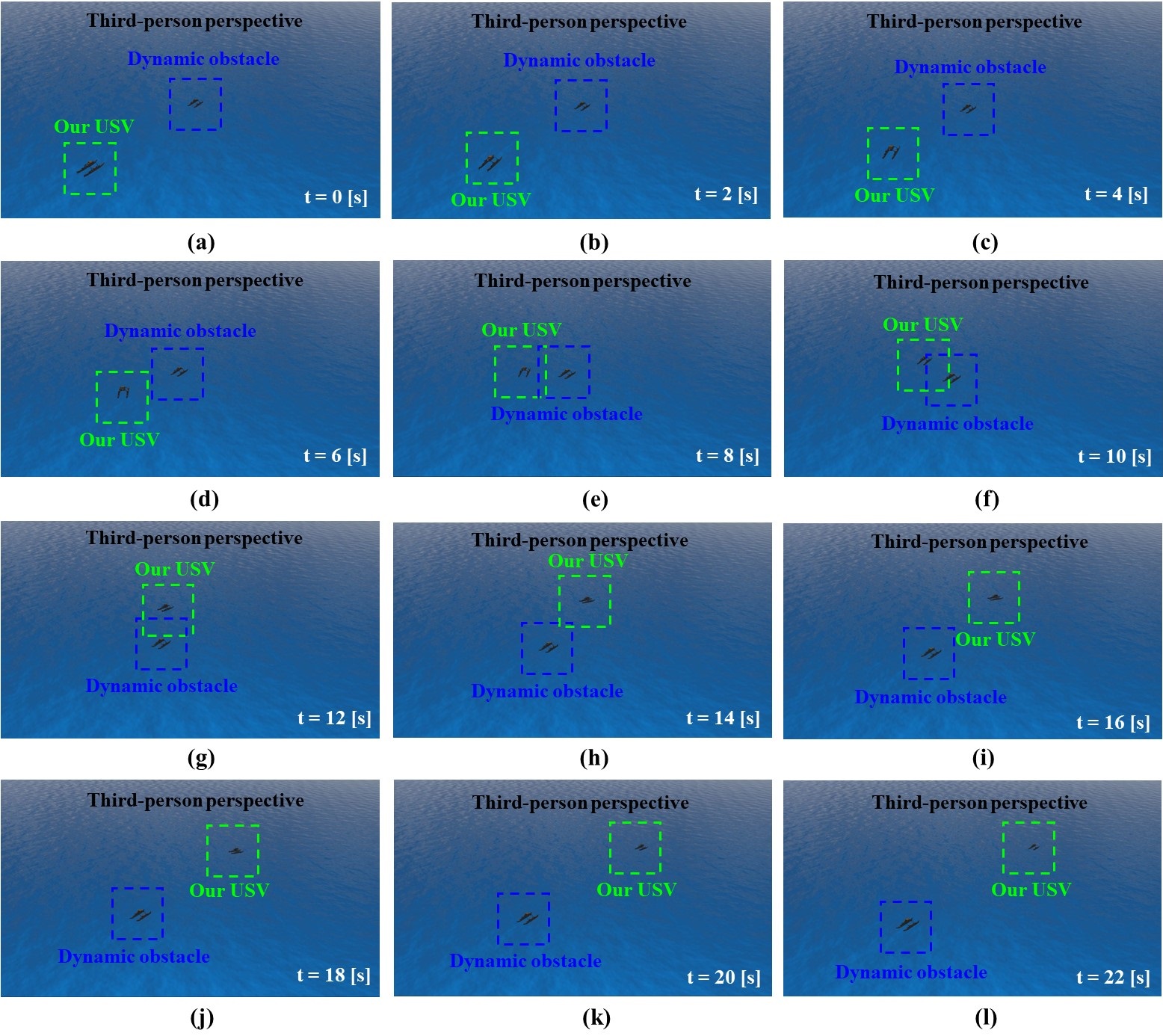}
 \centering
 \caption{The storyboards of the dynamic obstacle avoidance mission in a high-fidelity maritime environment in ROS Gazebo using the proposed D-GPMP2 algorithm. The start and goal positions of our USV are [50, 10] and [50, 90], respectively. The initial position of the dynamic obstacle is [50, 50]. From (a) to (l), the sub-figures demonstrate the positions of our USV and the dynamic obstacle when time equals to 0 [s], 2 [s], 4 [s], 6 [s], 8 [s], 10 [s], 12 [s], 14 [s], 16 [s], 18 [s], 20 [s] and 22 [s] from the third-person perspective. The high-fidelity maritime environment is provided by \cite{bingham2019toward}.}
 \label{ros_simulation} 
 \end{figure}

\subsection{Dynamic obstacle avoidance mission} 
 
 \hl{During the dynamic obstacle avoidance mission, the primary USV completes a package delivery mission between designated positions while evading a secondary USV moving at constant velocity.} We use the proposed D-GPMP2 algorithm to generate an optimised path based on the real-time status information of the dynamic obstacle when transporting the package. By following the optimised path generated by the proposed D-GPMP2 algorithm, \hl{the primary USV maintains successful dynamic obstacle avoidance throughout its entire transportation mission.} \hl{During the dynamic obstacle avoidance mission, it is assumed the USVs owns "perfect" autopilots, which indicates these USVs can follow the designated paths perfectly.} To provide a more intuitive understanding of the above process, the storyboards of the dynamic obstacle avoidance mission from the third-person perspectives is demonstrated in Fig. \ref{ros_simulation}.

\section{Conclusion}
\label{Conclusion}
 The work in this article successfully extended the traditional GP-based motion planning algorithm into complex and dynamic maritime environments. To achieve this purpose, a novel motion planning algorithm is proposed in this article, named as the D-GPMP2. By introducing dynamic obstacle area modelling and dynamic obstacle velocity modelling, we were able to use the proposed D-GPMP2 algorithm to generate safe, smooth and relatively short trajectories for USVs to avoid both static and dynamic obstacles simultaneously in maritime environments in an efficient manner. \hl{Furthermore, D-GPMP2-COLREGs is proposed, which is an enhanced variant of D-GPMP2 that incorporates COLREGs compliance for dynamic obstacle avoidance in maritime environments.} A variety of simulation environments across different platforms have been used to validate the practicality and functionality of the proposed algorithms. 
 
 \hl{More specifically, the contributions of this work can be summarised as follows:}
 \begin{itemize}
     \item \hl{This article first presented an approach for constructing adaptive safe areas for moving USVs. This is accomplished through integration of sensor measurements of dimensional parameters into the multivariate Gaussian distribution.}
     \item \hl{Then this article presented an approach for integrating the sensor measurements of the real-time motion state (speed, position and course) into the adaptive safe area. This is achieved through mathematical modelling innovations coupled with computer vision techniques based on the adaptive safe area.} 
     \item \hl{This article presented an approach for integrating COLREGs rules into the proposed motion planning algorithm, which further enhance its practicability in maritime environments.}
     \item \hl{This article also presented a new factor graph structure for GP-based motion planning, capable of handling both static and dynamic obstacles simultaneously.}
     \item \hl{Last, this article thoroughly tested and validated the proposed motion planning algorithms in both a series of self-constructed simulation environments and a high-fidelity maritime environment in ROS.} 
 \end{itemize}
 
 \hl{In terms of future research, some proposed areas of focus are summarised as: 1) extending the proposed D-GPMP2 algorithm to enable 3-D motion planning for AUVs and 2) validating the proposed D-GPMP2 algorithm through real-world USV experiments to further demonstrate its practicality.}

\bibliographystyle{elsarticle-num-names} 
\bibliography{Paper}

\begin{thebibliography}{47}
\expandafter\ifx\csname natexlab\endcsname\relax\def\natexlab#1{#1}\fi
\providecommand{\url}[1]{\texttt{#1}}
\providecommand{\href}[2]{#2}
\providecommand{\path}[1]{#1}
\providecommand{\DOIprefix}{doi:}
\providecommand{\ArXivprefix}{arXiv:}
\providecommand{\URLprefix}{URL: }
\providecommand{\Pubmedprefix}{pmid:}
\providecommand{\doi}[1]{\href{http://dx.doi.org/#1}{\path{#1}}}
\providecommand{\Pubmed}[1]{\href{pmid:#1}{\path{#1}}}
\providecommand{\bibinfo}[2]{#2}
\ifx\xfnm\relax \def\xfnm[#1]{\unskip,\space#1}\fi
\bibitem[{Meng et~al.(2022{\natexlab{a}})Meng, Liu, Bucknall, Guo, and Ji}]{meng2022anisotropic}
\bibinfo{author}{J.~Meng}, \bibinfo{author}{Y.~Liu}, \bibinfo{author}{R.~Bucknall}, \bibinfo{author}{W.~Guo}, \bibinfo{author}{Z.~Ji},
\newblock \bibinfo{title}{Anisotropic gpmp2: A fast continuous-time gaussian processes based motion planner for unmanned surface vehicles in environments with ocean currents},
\newblock \bibinfo{journal}{IEEE Transactions on Automation Science and Engineering} \bibinfo{volume}{19} (\bibinfo{year}{2022}{\natexlab{a}}) \bibinfo{pages}{3914--3931}.
\bibitem[{Meng et~al.(2022{\natexlab{b}})Meng, Humne, Bucknall, Englot, and Liu}]{meng2022fully}
\bibinfo{author}{J.~Meng}, \bibinfo{author}{A.~Humne}, \bibinfo{author}{R.~Bucknall}, \bibinfo{author}{B.~Englot}, \bibinfo{author}{Y.~Liu},
\newblock \bibinfo{title}{A fully-autonomous framework of unmanned surface vehicles in maritime environments using gaussian process motion planning},
\newblock \bibinfo{journal}{IEEE Journal of Oceanic Engineering} \bibinfo{volume}{48} (\bibinfo{year}{2022}{\natexlab{b}}) \bibinfo{pages}{59--79}.
\bibitem[{Dijkstra(2022)}]{dijkstra2022note}
\bibinfo{author}{E.~W. Dijkstra},
\newblock \bibinfo{title}{A note on two problems in connexion with graphs},
\newblock in: \bibinfo{booktitle}{Edsger Wybe Dijkstra: His Life, Work, and Legacy}, \bibinfo{year}{2022}, pp. \bibinfo{pages}{287--290}.
\bibitem[{Hart et~al.(1968)Hart, Nilsson, and Raphael}]{hart1968formal}
\bibinfo{author}{P.~E. Hart}, \bibinfo{author}{N.~J. Nilsson}, \bibinfo{author}{B.~Raphael},
\newblock \bibinfo{title}{A formal basis for the heuristic determination of minimum cost paths},
\newblock \bibinfo{journal}{IEEE transactions on Systems Science and Cybernetics} \bibinfo{volume}{4} (\bibinfo{year}{1968}) \bibinfo{pages}{100--107}.
\bibitem[{Petres et~al.(2007)Petres, Pailhas, Patron, Petillot, Evans, and Lane}]{petres2007path}
\bibinfo{author}{C.~Petres}, \bibinfo{author}{Y.~Pailhas}, \bibinfo{author}{P.~Patron}, \bibinfo{author}{Y.~Petillot}, \bibinfo{author}{J.~Evans}, \bibinfo{author}{D.~Lane},
\newblock \bibinfo{title}{Path planning for autonomous underwater vehicles},
\newblock \bibinfo{journal}{IEEE Transactions on Robotics} \bibinfo{volume}{23} (\bibinfo{year}{2007}) \bibinfo{pages}{331--341}.
\bibitem[{Kavraki et~al.(1996)Kavraki, Svestka, Latombe, and Overmars}]{kavraki1996probabilistic}
\bibinfo{author}{L.~E. Kavraki}, \bibinfo{author}{P.~Svestka}, \bibinfo{author}{J.-C. Latombe}, \bibinfo{author}{M.~H. Overmars},
\newblock \bibinfo{title}{Probabilistic roadmaps for path planning in high-dimensional configuration spaces},
\newblock \bibinfo{journal}{IEEE transactions on Robotics and Automation} \bibinfo{volume}{12} (\bibinfo{year}{1996}) \bibinfo{pages}{566--580}.
\bibitem[{LaValle(1998)}]{lavalle1998rapidly}
\bibinfo{author}{S.~LaValle},
\newblock \bibinfo{title}{Rapidly-exploring random trees: A new tool for path planning},
\newblock \bibinfo{journal}{Research Report 9811}  (\bibinfo{year}{1998}).
\bibitem[{Meng et~al.(2018)Meng, Pawar, Kay, and Li}]{meng2018uav}
\bibinfo{author}{J.~Meng}, \bibinfo{author}{V.~M. Pawar}, \bibinfo{author}{S.~Kay}, \bibinfo{author}{A.~Li},
\newblock \bibinfo{title}{Uav path planning system based on 3d informed rrt for dynamic obstacle avoidance},
\newblock in: \bibinfo{booktitle}{2018 IEEE International Conference on Robotics and Biomimetics (ROBIO)}, \bibinfo{organization}{IEEE}, \bibinfo{year}{2018}, pp. \bibinfo{pages}{1653--1658}.
\bibitem[{Gammell et~al.(2014)Gammell, Srinivasa, and Barfoot}]{gammell2014informed}
\bibinfo{author}{J.~D. Gammell}, \bibinfo{author}{S.~S. Srinivasa}, \bibinfo{author}{T.~D. Barfoot},
\newblock \bibinfo{title}{Informed rrt: Optimal sampling-based path planning focused via direct sampling of an admissible ellipsoidal heuristic},
\newblock in: \bibinfo{booktitle}{2014 IEEE/RSJ international conference on intelligent robots and systems}, \bibinfo{organization}{IEEE}, \bibinfo{year}{2014}, pp. \bibinfo{pages}{2997--3004}.
\bibitem[{Kuffner and LaValle(2000)}]{kuffner2000rrt}
\bibinfo{author}{J.~J. Kuffner}, \bibinfo{author}{S.~M. LaValle},
\newblock \bibinfo{title}{Rrt-connect: An efficient approach to single-query path planning},
\newblock in: \bibinfo{booktitle}{Proceedings 2000 ICRA. Millennium Conference. IEEE International Conference on Robotics and Automation. Symposia Proceedings (Cat. No. 00CH37065)}, volume~\bibinfo{volume}{2}, \bibinfo{organization}{IEEE}, \bibinfo{year}{2000}, pp. \bibinfo{pages}{995--1001}.
\bibitem[{Khatib(1986)}]{khatib1986real}
\bibinfo{author}{O.~Khatib},
\newblock \bibinfo{title}{Real-time obstacle avoidance for manipulators and mobile robots},
\newblock \bibinfo{journal}{The international journal of robotics research} \bibinfo{volume}{5} (\bibinfo{year}{1986}) \bibinfo{pages}{90--98}.
\bibitem[{Petres et~al.(2005)Petres, Pailhas, Petillot, and Lane}]{petres2005underwater}
\bibinfo{author}{C.~Petres}, \bibinfo{author}{Y.~Pailhas}, \bibinfo{author}{Y.~Petillot}, \bibinfo{author}{D.~Lane},
\newblock \bibinfo{title}{Underwater path planing using fast marching algorithms},
\newblock in: \bibinfo{booktitle}{Europe Oceans 2005}, volume~\bibinfo{volume}{2}, \bibinfo{organization}{IEEE}, \bibinfo{year}{2005}, pp. \bibinfo{pages}{814--819}.
\bibitem[{Colorni et~al.(1991)Colorni, Dorigo, Maniezzo et~al.}]{colorni1991distributed}
\bibinfo{author}{A.~Colorni}, \bibinfo{author}{M.~Dorigo}, \bibinfo{author}{V.~Maniezzo}, et~al.,
\newblock \bibinfo{title}{Distributed optimization by ant colonies},
\newblock in: \bibinfo{booktitle}{Proceedings of the first European conference on artificial life}, volume \bibinfo{volume}{142}, \bibinfo{organization}{Paris, France}, \bibinfo{year}{1991}, pp. \bibinfo{pages}{134--142}.
\bibitem[{Whitley(1994)}]{whitley1994genetic}
\bibinfo{author}{D.~Whitley},
\newblock \bibinfo{title}{A genetic algorithm tutorial},
\newblock \bibinfo{journal}{Statistics and computing} \bibinfo{volume}{4} (\bibinfo{year}{1994}) \bibinfo{pages}{65--85}.
\bibitem[{MahmoudZadeh et~al.(2016)MahmoudZadeh, Powers, and Yazdani}]{mahmoudzadeh2016novel}
\bibinfo{author}{S.~MahmoudZadeh}, \bibinfo{author}{D.~M. Powers}, \bibinfo{author}{A.~M. Yazdani},
\newblock \bibinfo{title}{A novel efficient task-assign route planning method for auv guidance in a dynamic cluttered environment},
\newblock in: \bibinfo{booktitle}{2016 IEEE Congress on Evolutionary Computation (CEC)}, \bibinfo{organization}{IEEE}, \bibinfo{year}{2016}, pp. \bibinfo{pages}{678--684}.
\bibitem[{Mukadam et~al.(2016)Mukadam, Yan, and Boots}]{mukadam2016gaussian}
\bibinfo{author}{M.~Mukadam}, \bibinfo{author}{X.~Yan}, \bibinfo{author}{B.~Boots},
\newblock \bibinfo{title}{Gaussian process motion planning},
\newblock in: \bibinfo{booktitle}{2016 IEEE international conference on robotics and automation (ICRA)}, \bibinfo{organization}{IEEE}, \bibinfo{year}{2016}, pp. \bibinfo{pages}{9--15}.
\bibitem[{Dong et~al.(2016)Dong, Mukadam, Dellaert, and Boots}]{dong2016motion}
\bibinfo{author}{J.~Dong}, \bibinfo{author}{M.~Mukadam}, \bibinfo{author}{F.~Dellaert}, \bibinfo{author}{B.~Boots},
\newblock \bibinfo{title}{Motion planning as probabilistic inference using gaussian processes and factor graphs.},
\newblock in: \bibinfo{booktitle}{Robotics: Science and Systems}, volume~\bibinfo{volume}{12}, \bibinfo{year}{2016}.
\bibitem[{Huang et~al.(2017)Huang, Mukadam, Liu, and Boots}]{huang2017motion}
\bibinfo{author}{E.~Huang}, \bibinfo{author}{M.~Mukadam}, \bibinfo{author}{Z.~Liu}, \bibinfo{author}{B.~Boots},
\newblock \bibinfo{title}{Motion planning with graph-based trajectories and gaussian process inference},
\newblock in: \bibinfo{booktitle}{2017 IEEE International Conference on Robotics and Automation (ICRA)}, \bibinfo{organization}{IEEE}, \bibinfo{year}{2017}, pp. \bibinfo{pages}{5591--5598}.
\bibitem[{Mukadam et~al.(2018)Mukadam, Dong, Yan, Dellaert, and Boots}]{mukadam2018continuous}
\bibinfo{author}{M.~Mukadam}, \bibinfo{author}{J.~Dong}, \bibinfo{author}{X.~Yan}, \bibinfo{author}{F.~Dellaert}, \bibinfo{author}{B.~Boots},
\newblock \bibinfo{title}{Continuous-time gaussian process motion planning via probabilistic inference},
\newblock \bibinfo{journal}{The International Journal of Robotics Research} \bibinfo{volume}{37} (\bibinfo{year}{2018}) \bibinfo{pages}{1319--1340}.
\bibitem[{Stentz(1994)}]{stentz1994optimal}
\bibinfo{author}{A.~Stentz},
\newblock \bibinfo{title}{Optimal and efficient path planning for partially-known environments},
\newblock in: \bibinfo{booktitle}{Proceedings of the 1994 IEEE international conference on robotics and automation}, \bibinfo{organization}{IEEE}, \bibinfo{year}{1994}, pp. \bibinfo{pages}{3310--3317}.
\bibitem[{Zhu et~al.(2021)Zhu, Yan, and Yue}]{zhu2021path}
\bibinfo{author}{X.~Zhu}, \bibinfo{author}{B.~Yan}, \bibinfo{author}{Y.~Yue},
\newblock \bibinfo{title}{Path planning and collision avoidance in unknown environments for usvs based on an improved d* lite},
\newblock \bibinfo{journal}{Applied Sciences} \bibinfo{volume}{11} (\bibinfo{year}{2021}) \bibinfo{pages}{7863}.
\bibitem[{Jiang et~al.(2023)Jiang, Zhu, and Xie}]{jiang2023dynamic}
\bibinfo{author}{C.~Jiang}, \bibinfo{author}{H.~Zhu}, \bibinfo{author}{Y.~Xie},
\newblock \bibinfo{title}{Dynamic obstacle avoidance research for mobile robots incorporating improved a-star algorithm and dwa algorithm},
\newblock in: \bibinfo{booktitle}{2023 3rd International Conference on Computer Science, Electronic Information Engineering and Intelligent Control Technology (CEI)}, \bibinfo{organization}{IEEE}, \bibinfo{year}{2023}, pp. \bibinfo{pages}{896--900}.
\bibitem[{He et~al.(2021)He, Chen, Chiang, and Li}]{he2021dynamic}
\bibinfo{author}{J.-H. He}, \bibinfo{author}{Y.-L. Chen}, \bibinfo{author}{H.-H. Chiang}, \bibinfo{author}{H.-C. Li},
\newblock \bibinfo{title}{Dynamic obstacle avoidance based on a dynamic space-time grid map for mobile robots},
\newblock in: \bibinfo{booktitle}{2021 IEEE International Conference on Consumer Electronics (ICCE)}, \bibinfo{organization}{IEEE}, \bibinfo{year}{2021}, pp. \bibinfo{pages}{1--3}.
\bibitem[{Enevoldsen et~al.(2021)Enevoldsen, Reinartz, and Galeazzi}]{enevoldsen2021colregs}
\bibinfo{author}{T.~T. Enevoldsen}, \bibinfo{author}{C.~Reinartz}, \bibinfo{author}{R.~Galeazzi},
\newblock \bibinfo{title}{Colregs-informed rrt* for collision avoidance of marine crafts},
\newblock in: \bibinfo{booktitle}{2021 IEEE International Conference on Robotics and Automation (ICRA)}, \bibinfo{organization}{IEEE}, \bibinfo{year}{2021}, pp. \bibinfo{pages}{8083--8089}.
\bibitem[{Chiang and Tapia(2018)}]{chiang2018colreg}
\bibinfo{author}{H.-T.~L. Chiang}, \bibinfo{author}{L.~Tapia},
\newblock \bibinfo{title}{Colreg-rrt: An rrt-based colregs-compliant motion planner for surface vehicle navigation},
\newblock \bibinfo{journal}{IEEE Robotics and Automation Letters} \bibinfo{volume}{3} (\bibinfo{year}{2018}) \bibinfo{pages}{2024--2031}.
\bibitem[{Ferguson et~al.(2006)Ferguson, Kalra, and Stentz}]{ferguson2006replanning}
\bibinfo{author}{D.~Ferguson}, \bibinfo{author}{N.~Kalra}, \bibinfo{author}{A.~Stentz},
\newblock \bibinfo{title}{Replanning with rrts},
\newblock in: \bibinfo{booktitle}{Proceedings 2006 IEEE International Conference on Robotics and Automation, 2006. ICRA 2006.}, \bibinfo{organization}{IEEE}, \bibinfo{year}{2006}, pp. \bibinfo{pages}{1243--1248}.
\bibitem[{Adiyatov and Varol(2017)}]{adiyatov2017novel}
\bibinfo{author}{O.~Adiyatov}, \bibinfo{author}{H.~A. Varol},
\newblock \bibinfo{title}{A novel rrt-based algorithm for motion planning in dynamic environments},
\newblock in: \bibinfo{booktitle}{2017 IEEE International Conference on Mechatronics and Automation (ICMA)}, \bibinfo{organization}{IEEE}, \bibinfo{year}{2017}, pp. \bibinfo{pages}{1416--1421}.
\bibitem[{Aoude et~al.(2013)Aoude, Luders, Joseph, Roy, and How}]{aoude2013probabilistically}
\bibinfo{author}{G.~S. Aoude}, \bibinfo{author}{B.~D. Luders}, \bibinfo{author}{J.~M. Joseph}, \bibinfo{author}{N.~Roy}, \bibinfo{author}{J.~P. How},
\newblock \bibinfo{title}{Probabilistically safe motion planning to avoid dynamic obstacles with uncertain motion patterns},
\newblock \bibinfo{journal}{Autonomous Robots} \bibinfo{volume}{35} (\bibinfo{year}{2013}) \bibinfo{pages}{51--76}.
\bibitem[{Falanga et~al.(2020)Falanga, Kleber, and Scaramuzza}]{falanga2020dynamic}
\bibinfo{author}{D.~Falanga}, \bibinfo{author}{K.~Kleber}, \bibinfo{author}{D.~Scaramuzza},
\newblock \bibinfo{title}{Dynamic obstacle avoidance for quadrotors with event cameras},
\newblock \bibinfo{journal}{Science Robotics} \bibinfo{volume}{5} (\bibinfo{year}{2020}) \bibinfo{pages}{eaaz9712}.
\bibitem[{Wu et~al.(2023)Wu, Zhang, Huang, Zhang, Luan, Zhao, and Chen}]{wu2023research}
\bibinfo{author}{H.~Wu}, \bibinfo{author}{Y.~Zhang}, \bibinfo{author}{L.~Huang}, \bibinfo{author}{J.~Zhang}, \bibinfo{author}{Z.~Luan}, \bibinfo{author}{W.~Zhao}, \bibinfo{author}{F.~Chen},
\newblock \bibinfo{title}{Research on vehicle obstacle avoidance path planning based on apf-pso},
\newblock \bibinfo{journal}{Proceedings of the Institution of Mechanical Engineers, Part D: Journal of Automobile Engineering} \bibinfo{volume}{237} (\bibinfo{year}{2023}) \bibinfo{pages}{1391--1405}.
\bibitem[{Sonny et~al.(2023)Sonny, Yeduri, and Cenkeramaddi}]{sonny2023q}
\bibinfo{author}{A.~Sonny}, \bibinfo{author}{S.~R. Yeduri}, \bibinfo{author}{L.~R. Cenkeramaddi},
\newblock \bibinfo{title}{Q-learning-based unmanned aerial vehicle path planning with dynamic obstacle avoidance},
\newblock \bibinfo{journal}{Applied Soft Computing} \bibinfo{volume}{147} (\bibinfo{year}{2023}) \bibinfo{pages}{110773}.
\bibitem[{Choi et~al.(2021)Choi, Lee, and Lee}]{choi2021reinforcement}
\bibinfo{author}{J.~Choi}, \bibinfo{author}{G.~Lee}, \bibinfo{author}{C.~Lee},
\newblock \bibinfo{title}{Reinforcement learning-based dynamic obstacle avoidance and integration of path planning},
\newblock \bibinfo{journal}{Intelligent Service Robotics} \bibinfo{volume}{14} (\bibinfo{year}{2021}) \bibinfo{pages}{663--677}.
\bibitem[{Chen et~al.(2019)Chen, Zhang, Zhang, Nie, Tang, and Zhu}]{chen2019hybrid}
\bibinfo{author}{Z.~Chen}, \bibinfo{author}{Y.~Zhang}, \bibinfo{author}{Y.~Zhang}, \bibinfo{author}{Y.~Nie}, \bibinfo{author}{J.~Tang}, \bibinfo{author}{S.~Zhu},
\newblock \bibinfo{title}{A hybrid path planning algorithm for unmanned surface vehicles in complex environment with dynamic obstacles},
\newblock \bibinfo{journal}{IEEE access} \bibinfo{volume}{7} (\bibinfo{year}{2019}) \bibinfo{pages}{126439--126449}.
\bibitem[{Qing et~al.(2023)Qing, Jiang, and Yin}]{qing2023dynamic}
\bibinfo{author}{W.~Qing}, \bibinfo{author}{P.~Jiang}, \bibinfo{author}{Y.~Yin},
\newblock \bibinfo{title}{Dynamic obstacle avoidance for uav motion planning in unknown environments based on probabilistic inference},
\newblock in: \bibinfo{booktitle}{2023 42nd Chinese Control Conference (CCC)}, \bibinfo{organization}{IEEE}, \bibinfo{year}{2023}, pp. \bibinfo{pages}{3822--3827}.
\bibitem[{Bingham et~al.(2019)Bingham, Ag{\"u}ero, McCarrin, Klamo, Malia, Allen, Lum, Rawson, and Waqar}]{bingham2019toward}
\bibinfo{author}{B.~Bingham}, \bibinfo{author}{C.~Ag{\"u}ero}, \bibinfo{author}{M.~McCarrin}, \bibinfo{author}{J.~Klamo}, \bibinfo{author}{J.~Malia}, \bibinfo{author}{K.~Allen}, \bibinfo{author}{T.~Lum}, \bibinfo{author}{M.~Rawson}, \bibinfo{author}{R.~Waqar},
\newblock \bibinfo{title}{Toward maritime robotic simulation in gazebo},
\newblock in: \bibinfo{booktitle}{OCEANS 2019 MTS/IEEE SEATTLE}, \bibinfo{organization}{IEEE}, \bibinfo{year}{2019}, pp. \bibinfo{pages}{1--10}.
\bibitem[{Barfoot et~al.(2014)Barfoot, Tong, and S{\"a}rkk{\"a}}]{barfoot2014batch}
\bibinfo{author}{T.~D. Barfoot}, \bibinfo{author}{C.~H. Tong}, \bibinfo{author}{S.~S{\"a}rkk{\"a}},
\newblock \bibinfo{title}{Batch continuous-time trajectory estimation as exactly sparse gaussian process regression.},
\newblock in: \bibinfo{booktitle}{Robotics: Science and Systems}, volume~\bibinfo{volume}{10}, \bibinfo{organization}{Citeseer}, \bibinfo{year}{2014}, pp. \bibinfo{pages}{1--10}.
\bibitem[{Bassett and Deride(2019)}]{bassett2019maximum}
\bibinfo{author}{R.~Bassett}, \bibinfo{author}{J.~Deride},
\newblock \bibinfo{title}{Maximum a posteriori estimators as a limit of bayes estimators},
\newblock \bibinfo{journal}{Mathematical Programming} \bibinfo{volume}{174} (\bibinfo{year}{2019}) \bibinfo{pages}{129--144}.
\bibitem[{Gauss(1877)}]{gauss1877theoria}
\bibinfo{author}{C.~F. Gauss}, \bibinfo{title}{Theoria motus corporum coelestium in sectionibus conicis solem ambientium}, volume~\bibinfo{volume}{7}, \bibinfo{publisher}{FA Perthes}, \bibinfo{year}{1877}.
\bibitem[{Levenberg(1944)}]{levenberg1944method}
\bibinfo{author}{K.~Levenberg},
\newblock \bibinfo{title}{A method for the solution of certain non-linear problems in least squares},
\newblock \bibinfo{journal}{Quarterly of applied mathematics} \bibinfo{volume}{2} (\bibinfo{year}{1944}) \bibinfo{pages}{164--168}.
\bibitem[{opt(2025)}]{opt2025}
\bibinfo{title}{Ocean power technologies}, \bibinfo{year}{2025}. \URLprefix \url{https://oceanpowertechnologies.com/}.
\bibitem[{Xiao et~al.(2015)Xiao, Ligteringen, Van~Gulijk, and Ale}]{xiao2015comparison}
\bibinfo{author}{F.~Xiao}, \bibinfo{author}{H.~Ligteringen}, \bibinfo{author}{C.~Van~Gulijk}, \bibinfo{author}{B.~Ale},
\newblock \bibinfo{title}{Comparison study on ais data of ship traffic behavior},
\newblock \bibinfo{journal}{Ocean Engineering} \bibinfo{volume}{95} (\bibinfo{year}{2015}) \bibinfo{pages}{84--93}.
\bibitem[{Andrade(2020)}]{andrade2020development}
\bibinfo{author}{E.~d. Andrade},
\newblock \bibinfo{title}{Development of an autonomous propulsion system for surface vessels},
\newblock \bibinfo{journal}{Rio de Janeiro, Brazil: Federal University of Rio de Janeiro. http://www. monografias. poli. ufrj. br/mon ografias/monopoli10032050. pdf}  (\bibinfo{year}{2020}).
\bibitem[{{H}{E}{R}{O}{N}(2023)}]{heron}
\bibinfo{author}{{H}{E}{R}{O}{N}}, \bibinfo{year}{2023}. \URLprefix \url{{https://geo-matching.com/usvs-unmanned-surface-vehicles/heron-usv}}.
\bibitem[{Jwo et~al.(2023)Jwo, Cho, and Biswal}]{jwo2023geometric}
\bibinfo{author}{D.-J. Jwo}, \bibinfo{author}{T.-S. Cho}, \bibinfo{author}{A.~Biswal},
\newblock \bibinfo{title}{Geometric insights into the multivariate gaussian distribution and its entropy and mutual information},
\newblock \bibinfo{journal}{Entropy} \bibinfo{volume}{25} (\bibinfo{year}{2023}) \bibinfo{pages}{1177}.
\bibitem[{Prince(2012)}]{prince2012computer}
\bibinfo{author}{S.~J. Prince}, \bibinfo{title}{Computer vision: models, learning, and inference}, \bibinfo{publisher}{Cambridge University Press}, \bibinfo{year}{2012}.
\bibitem[{col(2025)}]{colreg}
\bibinfo{title}{Colreg - preventing collisions at sea}, \bibinfo{year}{2025}. \URLprefix \url{https://www.imo.org/en/OurWork/Safety/Pages/Preventing-Collisions.aspx}.
\bibitem[{{G}oogle~{E}arth 10.38.0.0(2021)}]{googleearth}
\bibinfo{author}{{G}oogle~{E}arth 10.38.0.0}, \bibinfo{title}{{R}oadfold {L}ake 50°42'08"{N} 4°13'43"{W}}, \bibinfo{year}{2021}. \URLprefix \url{{https://earth.google.com/web/@50.70225035,-4.22869228,125.09519554a,5902.19941083d,35y,99.68055535h,0t,0r/data=OgMKATA}}, \bibinfo{note}{[Accessed 15-10-2023]}.

\end{thebibliography}

\end{document}